\documentclass[10pt,twocolumn,letterpaper]{article}

\usepackage[pagenumbers]{cvpr} 

\usepackage[accsupp]{axessibility} 

\usepackage{amsmath,amsfonts,amssymb}
\usepackage{multirow}
\usepackage{makecell}
\usepackage{float}
\usepackage{stfloats}
\usepackage{algorithm}

\definecolor{cvprblue}{rgb}{0.21,0.49,0.74}
\usepackage[pagebackref,breaklinks,colorlinks,allcolors=cvprblue]{hyperref}

\def\paperID{*****} 
\def\confName{CVPR}
\def\confYear{2026}

\title{MambaLiteUNet: Cross-Gated Adaptive Feature Fusion\\for Robust Skin Lesion Segmentation}

\author{
Md Maklachur Rahman$^{1}$ \quad
Soon Ki Jung$^{2}$ \quad
Tracy Hammond$^{1}$\\
$^{1}$Texas A\&M University, College Station, TX, USA\\
$^{2}$Kyungpook National University, Daegu, South Korea\\
{\tt\small \{maklachur,hammond\}@tamu.edu \quad skjung@knu.ac.kr}
}

\begin{document}

\maketitle
\begin{abstract}
Recent segmentation models have demonstrated promising efficiency by aggressively reducing parameter counts and computational complexity. However, these models often struggle to accurately delineate fine lesion boundaries and texture patterns essential for early skin cancer diagnosis and treatment planning. In this paper, we propose MambaLiteUNet, a compact yet robust segmentation framework that integrates Mamba state space modeling into a U-Net architecture, along with three key modules: Adaptive Multi-Branch Mamba Feature Fusion (AMF), Local-Global Feature Mixing (LGFM), and Cross-Gated Attention (CGA). These modules are designed to enhance local–global feature interaction, preserve spatial details, and improve the quality of skip connections. MambaLiteUNet achieves an average IoU of 87.12\% and average Dice score of 93.09\% across ISIC2017, ISIC2018, HAM10000, and PH2 benchmarks, outperforming state-of-the-art models.
Compared to U-Net, our model improves average IoU and Dice by 7.72 and 4.61 points, respectively, while reducing parameters by 93.6\% and GFLOPs by 97.6\%. 
Additionally, in domain generalization with six unseen lesion categories, MambaLiteUNet achieves 77.61\% IoU and 87.23\% Dice, performing best among all evaluated models. Our extensive experiments demonstrate that MambaLiteUNet achieves a strong balance between accuracy and efficiency, making it a competitive and practical solution for dermatological image segmentation.
Our code is publicly available at: https://github.com/maklachur/MambaLiteUNet.
\end{abstract}

\begin{figure*}[t]
    \centering
    \includegraphics[width=0.87\textwidth]{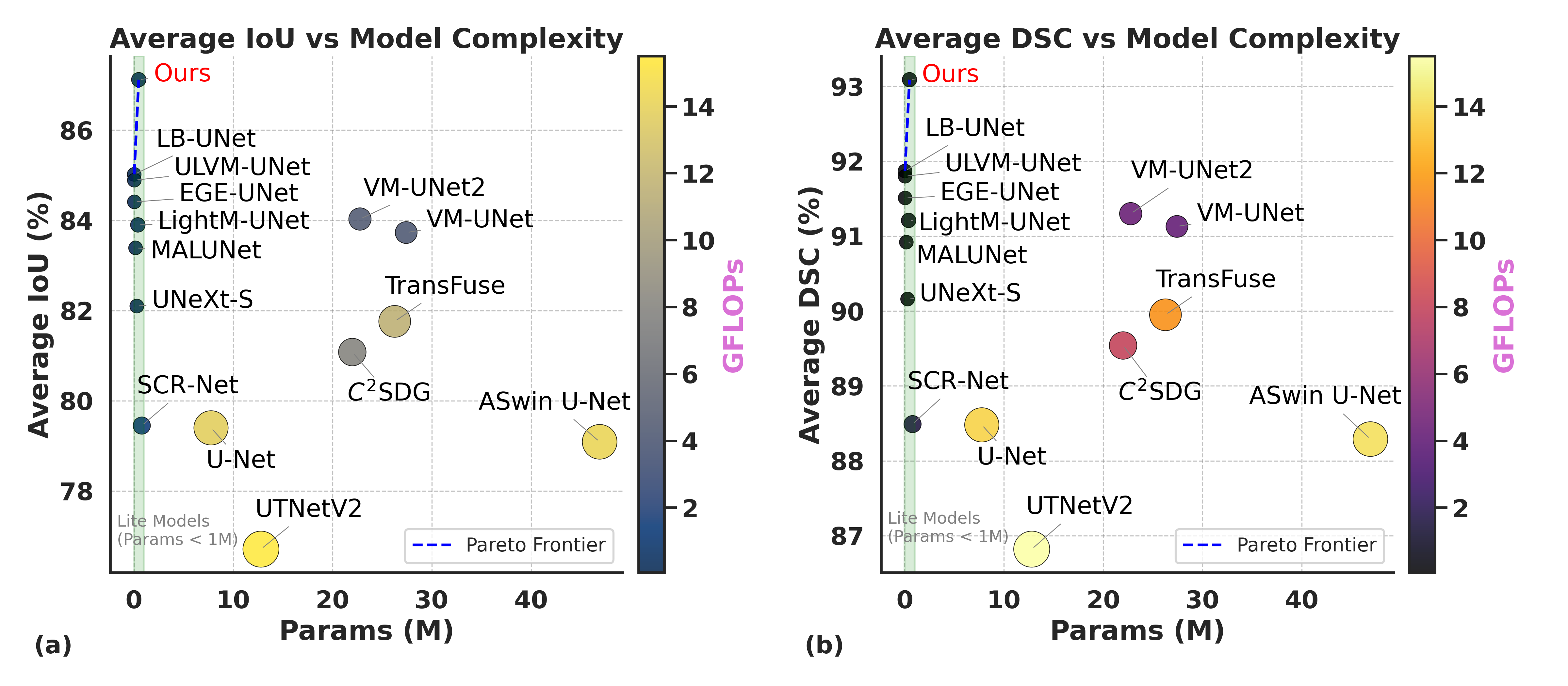}
    \caption{Complexity–performance trade‑off of SOTA segmentation models based on average IoU and average DSC across ISIC2017, ISIC2018, HAM10000, and PH2 datasets.}
    \label{fig:tradeoff_plot}
\end{figure*}

\section{Introduction}
Skin lesion segmentation is a fundamental task in computer-aided dermatological diagnosis, important for early cancer detection \cite{esteva2017dermatologist}, treatment planning \cite{ham10000}, and disease monitoring \cite{melanoma}. Precise lesion boundary delineation is especially important because subtle irregularities in shape, size, and extent may indicate malignancy, particularly in early-stage melanoma \cite{surveyskinlegion}. However, accurate segmentation remains challenging due to low contrast between lesions and surrounding skin, large variations in lesion appearance and morphology \cite{ham10000}, and imaging artifacts such as hair occlusions and specular reflections \cite{surveyskinlegion}. These difficulties become even more critical in mobile and point-of-care settings, where segmentation models must maintain strong accuracy under limited computational resources.

Encoder-decoder convolutional networks \cite{unet,unet++,aulunet,egeunet,lbunet, attentionunet} have long been the popular choice for medical image segmentation because they efficiently capture local textures and support dense prediction. However, convolution-based models struggle to model long-range spatial dependencies, which are important for lesions with irregular boundaries, disconnected regions, or ambiguous appearance. Transformer-based models improve global context modeling, but their quadratic complexity leads to substantial memory and computation overhead, which makes them less suitable for lightweight deployment.

Recently, State Space Models (SSMs), particularly Mamba-based vision architectures, have emerged as a promising alternative for modeling long-range dependencies with linear-time complexity \cite{mambaoriginal,visionmambaEfficient,vmamba}. This makes them promising for dense prediction tasks that require both efficiency and broad contextual reasoning. However, existing Mamba-based segmentation models often rely on static feature fusion and conventional skip connections, which can limit multi-scale representation learning and weaken boundary refinement in challenging lesion regions.

We address these limitations by introducing MambaLiteUNet, a lightweight yet robust skin lesion segmentation framework that combines Vision Mamba state space layers \cite{vmamba} with a compact U-Net backbone \cite{unet, aulunet}. We introduce three novel modules to enhance the overall discriminative ability of our model, which helps improve lesion segmentation performance without adding excessive computational complexity (Figure~\ref{fig:tradeoff_plot}). 
Specifically, Adaptive Multi-Branch Mamba Feature Fusion (AMF) dynamically partitions feature channels into parallel state space branches and employs dual gating mechanisms to enhance multi-scale feature representation. Local-Global Feature Mixing (LGFM) jointly leverages depthwise convolutions and multi-head self-attention to precisely capture both local texture details and global contextual information. Cross-Gated Attention (CGA) selectively filters and refines skip connections between encoder and decoder paths, significantly improving boundary delineation and spatial consistency.

In summary, our key contributions are as follows:

\begin{itemize}
    \item We propose MambaLiteUNet, a robust segmentation framework that effectively integrates Vision Mamba state-space modeling into a lightweight U-Net pipeline.
    
    \item We introduce AMF, LGFM, and CGA modules to improve multi-scale feature aggregation, enhance local and global context fusion, and refine encoder–decoder interactions for accurate boundary delineation.
    
    \item Extensive experiments on ISIC2017 \cite{isic2017challenge}, ISIC2018 \cite{isic2018challenge}, HAM10000 \cite{ham10000}, and PH2 \cite{ph2dataset} show MambaLiteUNet achieves top performance across all datasets, with an average of 93.09\% Dice and 87.12\% IoU, outperforming recent state-of-the-art (SOTA) models. In a domain generalization test on six unseen lesion types (AKIEC, BCC, BKL, DF, MEL, VASC) of HAM10000, it demonstrates strong generalization, with 87.23\% Dice and 77.61\% IoU, confirming its robustness.
    
\end{itemize}

\section{Related Work}
Medical image segmentation has been widely studied using encoder--decoder architectures, with U-Net \cite{unet} and its variants \cite{unet++,attentionunet,scrnet,malunet,egeunet,lbunet} forming the main foundation. These CNN-based models are efficient and effective at capturing local lesion appearance, but their locality often limits long-range spatial reasoning, which is important for lesions with irregular boundaries, disconnected regions, or low contrast. Transformer-based methods, such as TransUNet \cite{transunet}, TransFuse \cite{transfuse}, UTNetV2 \cite{utnetv2}, ASwin U-Net \cite{attentionswinunet}, and DS-TransUNet \cite{dstransunet}, improve global context modeling through self-attention. However, their quadratic complexity leads to higher memory and computational cost, which reduces their practicality in resource-constrained settings.

Recently, SSMs, especially Mamba-based vision architectures, have emerged as an efficient alternative for modeling long-range dependencies with linear-time complexity \cite{mambaoriginal,visionmambaEfficient,vmamba,mambainvision}. Vision Mamba \cite{visionmambaEfficient} and VMamba \cite{vmamba} show that state-space modeling can be effectively adapted to visual tasks while preserving scalability. Following this line of work, segmentation methods such as VM-UNet \cite{vmunet,vmunet2}, VM-UNet2 \cite{vmunet2}, LightM-UNet \cite{lightmunet}, and ULVM-UNet \cite{ultralightvmnet} incorporate Mamba-style modules for skin lesion segmentation. These methods improve the efficiency--accuracy trade-off, but many still rely on fixed feature fusion and standard skip connections, which can limit multi-scale interaction and boundary refinement.

In parallel, lightweight segmentation models have been developed to reduce parameters and FLOPs for practical deployment, including UNeXt-S \cite{unext}, SCR-Net \cite{scrnet}, MALUNet \cite{malunet}, EGE-UNet \cite{egeunet}, LB-UNet \cite{lbunet}, LightM-UNet \cite{lightmunet}, ULVM-UNet \cite{ultralightvmnet}, and AULUNet \cite{aulunet}. Although these models are computationally efficient, a compact design can come with weaker multi-scale representation learning and less precise boundary recovery under challenging lesion variations. In contrast, MambaLiteUNet combines Vision Mamba modeling with adaptive multi-branch fusion~\cite{resnext}, local-global feature mixing, and cross-gated skip refinement, which improves lesion representation and boundary delineation while maintaining low computational cost.
\section{Methodology}
\label{sec:methodology}
\begin{figure*}[t]
    \centering
    \includegraphics[width=0.98\textwidth]{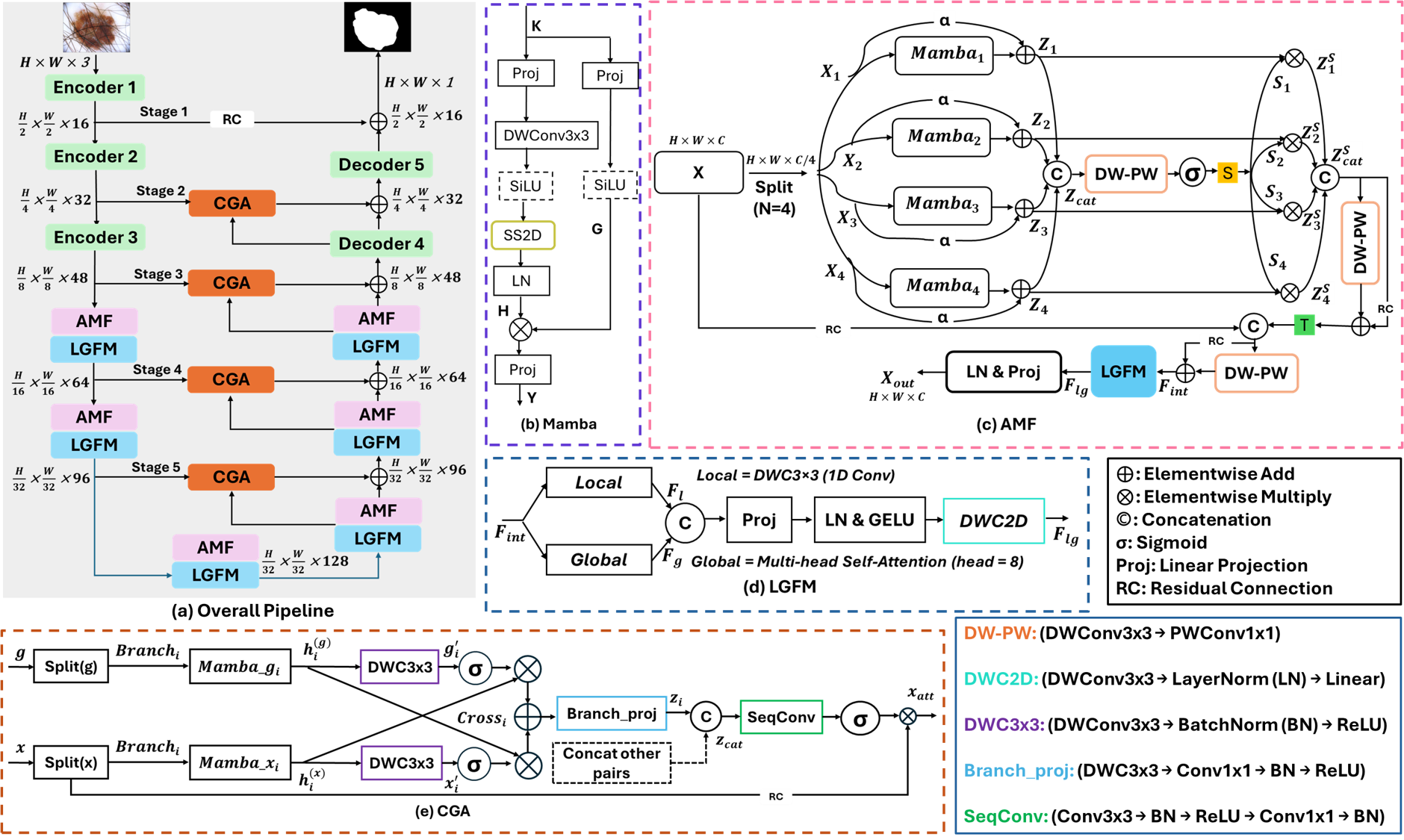}
    \caption{(a) Overall MambaLiteUNet pipeline. (b) Mamba block: Integrate SS2D \cite{vmamba}, leveraging SSM \cite{mambaoriginal} to process 2D visual data. (c) AMF: Merge parallel Mamba branches via learnable dual adaptive gating \( S \) and \( T \). (d) LGFM: Fuse local features (depthwise convolutions) with global context (multi-head attention). (e) CGA: Enhance skip connections using encoder-decoder features.}
    \label{fig:main_pipeline}
\end{figure*}

\subsection{Overall Architecture}
\label{subsec:architecture}
The overall pipeline of our model is shown in Figure~\ref{fig:main_pipeline}(a). It follows a U-Net-inspired design~\cite{unet,unet++} with five stages and a bottleneck. The encoder increases channels and the decoder gradually reduces them, using channel capacities of $\{16,32,48,64,96,128\}$. The early stages use standard convolutions with Group Normalization, while the deeper stages incorporate AMF and LGFM for stronger feature learning. Downsampling is performed by max pooling. Each skip connection is refined by CGA before fusion with decoder features. Finally, a $1\times1$ convolution followed by a sigmoid activation produces the lesion probability map.

\subsection{Mamba Block for State Space Modeling}
\label{subsec:mamba}
We derive our Mamba block (Figure \ref{fig:main_pipeline}(b)) from VMamba~\cite{vmamba},  forming the core of our AMF and CGA modules, as shown in Figure \ref{fig:main_pipeline}(c) and \ref{fig:main_pipeline}(e), respectively. Starting with layer normalized tokens \(K\in\mathbb{R}^{B\times N\times C}\) (where \(N=H\times W\)), we first compute a gating map \(G=\mathrm{SiLU}(K\,W_{g})\), with a learnable projection \(W_{g}\in\mathbb{R}^{C\times C}\) and \(\mathrm{SiLU}\) activation \cite{silu}. In parallel, we apply a second learnable projection \(W_{z}\in\mathbb{R}^{C\times C}\), followed by SiLU, a $3\times3$ depthwise convolution, and the SS2D module \cite{vmamba, mambaoriginal}. SS2D scans tokens in four directions, processes each sequence with an independent S6 block \cite{mambaoriginal}, and aggregates the outputs into a global context feature. We then apply layer normalization (LN) to obtain \(H\). Finally, we fuse local and global information by elementwise product \(Y=G\odot H\), generating transformer‑level receptive fields in linear time.

\subsection{Adaptive Multi-Branch Mamba Feature Fusion (AMF)}
Our AMF module (Figure \ref{fig:main_pipeline}(c)) enhances representational richness by unifying diverse feature streams through parallel Mamba‑based SSM blocks and dynamic gating. Given an input tensor $X\in\mathbb{R}^{B\times C\times H\times W}$,
we first split its channels into four equal groups 
\(\{X_k\in\mathbb{R}^{B\times \tfrac{C}{4}\times H\times W}\}_{k=1}^4\). 
Each group passes through a Mamba block that captures long‑range dependencies via a state‑space formulation. We add a scaled residual to preserve low‑level features:
\begin{equation}
    Z_k = \mathrm{Mamba}_k(X_k)\;+\;\alpha\,X_k,\quad k=1\dots4.
\end{equation}

Here, \(\alpha\) is a trainable scalar initialized to 0, enabling the network to progressively modulate the residual contribution over training.
Concatenating \(\{Z_k\}\) along the channel axis yields 
\(\,Z_{\mathrm{cat}}\in\mathbb{R}^{B\times C\times H\times W}\), 
ensuring sensitivity to both fine structures and global context.

To adaptively emphasize the most informative channels, we apply a two‑stage gating pipeline. In the first spatial (S) stage, we compute
\begin{equation}
S = \sigma\!\Bigl(\mathrm{PW}\bigl(\mathrm{DW}_{3\times3}(Z_{\mathrm{cat}})\bigr)\Bigr),
\end{equation}
where \(\mathrm{DW}_{3\times3}\) denotes a $3\times3$ depth‑wise convolution and \(\mathrm{PW}\) a point‑wise convolution. The sigmoid activation \(\sigma\) produces a gating tensor \(S\in\mathbb{R}^{B\times C\times H\times W}\). We then scale each branch $S_k$ by elementwise product $\odot$ with $Z_k$ as:
\begin{equation}
    Z_k^{S} = S_k \odot Z_k,\quad k=1\dots4,
\end{equation}
and concatenate \(\{Z_k^S\}\) into \(Z_{\mathrm{cat}}^{S}\). In the second transform (T) stage, \(Z_{\mathrm{cat}}^{S}\) is further refined by another pair of depth‑wise and point‑wise convolutions with a residual shortcut to yield gated tensor \(T\). Finally, we fuse it with the original input  
($F_{\mathrm{int}} = T + X$)
and forward it to the LGFM module (Section~\ref{sec:lgfm}). This design enables AMF to learn per-channel importance in a lightweight, multi-branch way.

\subsection{Local‑Global Feature Mixing (LGFM)}
\label{sec:lgfm}
LGFM, shown in Figure \ref{fig:main_pipeline}(d), refines the intermediate feature $F_{\mathrm{int}}\in\mathbb{R}^{B\times C\times H\times W}$ by integrating fine‑grained details with global context. We first extract local patterns $F_{\ell}$ using a $3\times3$ depthwise convolution.
In parallel, we capture global dependencies via multi‑head self‑attention \cite{transformer,vit} with eight heads \(h=8\), ensuring at each stage that \(C\) is divisible by \(h\) (so head\_dim \(=C/h\)). We flatten spatial dimensions to tokens \(N=H\times W\), project to queries, keys, and values, apply attention across \(h\) heads, then reshape back to \(\mathbb{R}^{B\times C\times H\times W}\) to obtain \(F_{g}\). We fuse these two paths into $F_{\mathrm{\ell g}}$ by channel‑concatenation and projection:
\begin{equation}
    F_{\mathrm{\ell g}} = \mathrm{DW}_{3\times3}\Bigl(\mathrm{GELU}\bigl(\mathrm{LN}\bigl(\mathrm{Conv}_{1\times1}([F_{\ell},F_{g}])\bigr)\bigr)\Bigr),
\end{equation}
where \(\mathrm{Conv}_{1\times1}\) reduces \(2C\) channels back to \(C\), \(\mathrm{LN}\) denotes LayerNorm, and GELU \cite{gelu} adds nonlinearity. 
This dual-path preserves lesion textures and integrates long-range features critical for accurate boundary delineation.

\subsection{Cross‑Gated Attention (CGA)}
We enhance skip connections with a CGA mechanism, shown in Figure \ref{fig:main_pipeline}(e), that adaptively filters and fuses encoder and decoder features. Let 
$x,\;g\;\in\;\mathbb{R}^{B\times C\times H\times W}$
denote encoder and decoder feature maps, respectively. We divide each into four pairs \(\{x_i,g_i\}_{i=1}^4\). Each pair is refined via a Mamba block, producing \(h_i^{(x)}\) and \(h_i^{(g)}\). We then pass it through a \(3\times3\) depth-wise convolution to derive $g'_i$ and $x'_i$, and then compute a pairwise cross-gating as follows:
\begin{equation}
    \mathrm{cross}_i 
= h_i^{(x)} \odot \sigma\bigl(g'_i\bigr)
+ h_i^{(g)} \odot \sigma\bigl(x'_i\bigr),
\end{equation}
where \(\sigma\) is sigmoid. Similarly, we compute the remaining pairs' gating and concatenate them to obtain \(Z_{\mathrm{cat}}\). 

Using \(Z_{\mathrm{cat}}\), we generate an attention mask \(\psi\) and apply it to the encoder feature \(x\) as follows:
\begin{gather}
  \psi = \sigma\bigl(\mathrm{Conv}_{3\times3}(\mathrm{ReLU}(\mathrm{BN}(Z_{\mathrm{cat}})))\bigr),\\
x_{\mathrm{att}} = \psi \odot x.  
\end{gather}

We then pass the attentional feature to the next decoder. The gated skip connection learns discriminative features by assigning more weight to foreground and less to background features. Therefore, it reduces background noise and emphasizes important lesion structures before they combine with decoder outputs in the next stage.

\section{Experiments and Results}
\newcommand{\meansd}[2]{#1\textsuperscript{#2}}

\subsection{Datasets, Implementations, and Evaluations}
We evaluate our model on four widely used skin lesion segmentation benchmarks: ISIC2017 \cite{isic2017challenge}, ISIC2018 \cite{isic2018challenge}, HAM10000 \cite{ham10000}, and PH2 \cite{ph2dataset}. ISIC2017 and ISIC2018 were curated by the International Skin Imaging Collaboration (ISIC) and provide high-quality dermoscopic images with lesion masks. HAM10000 is a diverse dataset with over 10,000 dermatoscopic images across seven lesion categories: melanocytic nevi (NV), melanoma (MEL), benign keratosis-like lesions (BKL), basal cell carcinoma (BCC), actinic keratoses and intraepithelial carcinoma (AKIEC), vascular lesions (VASC), and dermatofibroma (DF). PH2 is a smaller dataset of 200 dermoscopic images, mainly focused on melanocytic lesions with expert-annotated masks.

Following prior works \cite{ultralightvmnet,mucmnet,vmunet,lightmunet}, we adopt the training, validation, and test splits for ISIC2017 and ISIC2018. For HAM10000, we follow the dataset preparation strategy in \cite{matchseg}. For the cross-domain evaluation setting, we follow MatchSeg \cite{matchseg} and train the model exclusively on the dominant NV (melanocytic nevi) class, testing its ability to generalize to the six unseen lesion types (MEL, BKL, BCC, AKIEC, VASC, and DF). This setting reflects a realistic clinical scenario where labeled data is available for common benign cases, but pixel-level annotations are scarce for rarer or malignant lesions. For PH2, we split the data into train--val--test sets in a $7:1:2$ ratio. For all datasets, the input images are normalized, resized to $256\times256$, and augmented to improve the robustness.

We conduct all experiments on a single NVIDIA GeForce RTX 3090 Ti GPU with 24 GB of VRAM. We train our model using a combination of binary cross-entropy (BCE) and Dice loss (Dice) \cite{losssurvey}, defined as:

\begin{gather}
L_{\mathrm{BCE}} = -\frac{1}{N}\sum_{i=1}^N\bigl(g_i\log p_i + (1-g_i)\log(1-p_i)\bigr), \\
L_{\mathrm{Dice}} = 1 - \frac{2\times|X\cap Y|}{|X| + |Y|}, \\
L_{\mathrm{Total}} = L_{\mathrm{BCE}} + L_{\mathrm{Dice}},
\end{gather}
where $N$ denotes the number of samples, $g_i$ and $p_i$ are the ground truth and predicted probabilities for pixel $i$, and $X$, $Y$ represent the sizes of the ground truth and predicted masks. We train for 300 epochs using AdamW \cite{adamw} with an initial learning rate of 0.001, decayed to 0.00001 via cosine annealing \cite{cosineannealing}, and use a batch size of 8.

To evaluate, following \cite{ultralightvmnet,mucmnet,matchseg}, we report Intersection over Union (IoU), Dice Similarity Coefficient (DSC), Accuracy (AC), Sensitivity (SE), Specificity (SP), and the 95th percentile Hausdorff Distance (HD95). Equations and HD95 results are given in Supp. Secs. 6 and 7, respectively. We also report parameter count (M) and GFLOPs for a $256\times256$ input.


\begin{table*}[t]
    \centering
    \footnotesize
    \setlength{\tabcolsep}{.4mm}
    \renewcommand{\arraystretch}{1.14}
    \begin{tabular}{l|c|c|c|c|c||c|c|c|c|c||c|c|c|c|c}
        \hline 
        \multirow{2}{*}{Model} 
        & \multicolumn{5}{c||}{ISIC2017}
        & \multicolumn{5}{c||}{ISIC2018} 
        & \multicolumn{5}{c}{HAM10000} \\
        \cline{2-16}
        & IoU$\uparrow$ & DSC$\uparrow$ & AC$\uparrow$ & SP$\uparrow$ & SE$\uparrow$
        & IoU$\uparrow$ & DSC$\uparrow$ & AC$\uparrow$ & SP$\uparrow$ & SE$\uparrow$
        & IoU$\uparrow$ & DSC$\uparrow$ & AC$\uparrow$ & SP$\uparrow$ & SE$\uparrow$ \\
        \hline 
        U-Net \cite{unet}
        & \meansd{79.55}{.24} & \meansd{88.61}{.26} & \meansd{95.72}{.20} & \meansd{97.43}{.22} & \meansd{88.36}{.64}
        & \meansd{74.64}{.25} & \meansd{85.48}{.28} & \meansd{94.18}{.18} & \meansd{97.66}{.24} & \meansd{81.17}{.69}
        & \meansd{83.07}{.28} & \meansd{90.75}{.30} & \meansd{95.21}{.19} & \meansd{96.02}{.26} & \meansd{92.84}{.71} \\

        SCR-Net \cite{scrnet} 
        & \meansd{78.57}{.28} & \meansd{88.00}{.31} & \meansd{95.60}{.21} & \meansd{97.93}{.24} & \meansd{85.57}{.78}
        & \meansd{79.27}{.46} & \meansd{88.44}{.42} & \meansd{95.06}{.20} & \meansd{96.54}{.27} & \meansd{89.52}{.82}
        & \meansd{85.86}{.40} & \meansd{92.39}{.37} & \meansd{96.09}{.22} & \meansd{96.85}{.29} & \meansd{93.86}{.77} \\

        TransFuse \cite{transfuse}
        & \meansd{80.17}{.31} & \meansd{89.00}{.29} & \meansd{95.94}{.18} & \meansd{97.98}{.26} & \meansd{87.14}{.70}
        & \meansd{78.75}{.36} & \meansd{88.11}{.33} & \meansd{94.80}{.19} & \meansd{95.74}{.28} & \meansd{\textbf{91.28}}{.73}
        & \meansd{84.59}{.34} & \meansd{91.65}{.32} & \meansd{95.53}{.21} & \meansd{95.02}{.27} & \meansd{97.03}{.66} \\

        UTNetV2 \cite{utnetv2}
        & \meansd{78.35}{.19} & \meansd{87.86}{.22} & \meansd{95.54}{.23} & \meansd{98.05}{.27} & \meansd{84.85}{.82}
        & \meansd{77.46}{.31} & \meansd{87.30}{.29} & \meansd{94.60}{.21} & \meansd{96.48}{.25} & \meansd{87.60}{.76}
        & \meansd{75.82}{.26} & \meansd{86.25}{.25} & \meansd{92.32}{.24} & \meansd{91.33}{.31} & \meansd{95.24}{.68} \\

        ASwin U-Net \cite{attentionswinunet}
        & \meansd{78.37}{.23} & \meansd{87.87}{.27} & \meansd{95.53}{.19} & \meansd{97.87}{.25} & \meansd{85.52}{.75}
        & \meansd{74.62}{.24} & \meansd{85.46}{.27} & \meansd{94.19}{.22} & \meansd{97.66}{.26} & \meansd{81.17}{.71}
        & \meansd{81.96}{.33} & \meansd{90.09}{.28} & \meansd{94.56}{.23} & \meansd{93.52}{.30} & \meansd{\textbf{97.66}}{.59} \\

        $C^2$SDG \cite{c2sdg}
        & \meansd{80.73}{.40} & \meansd{89.34}{.38} & \meansd{96.01}{.17} & \meansd{97.72}{.23} & \meansd{88.65}{.66}
        & \meansd{80.00}{.45} & \meansd{88.88}{.43} & \meansd{95.22}{.19} & \meansd{96.53}{.24} & \meansd{90.32}{.70}
        & \meansd{81.34}{.29} & \meansd{89.71}{.27} & \meansd{94.37}{.21} & \meansd{93.48}{.28} & \meansd{97.00}{.62} \\

        UNeXt-S \cite{unext}
        & \meansd{80.91}{.37} & \meansd{89.45}{.35} & \meansd{96.06}{.18} & \meansd{\textbf{98.06}}{.29} & \meansd{87.58}{.74}
        & \meansd{80.29}{.52} & \meansd{89.07}{.50} & \meansd{95.57}{.20} & \meansd{\textbf{98.18}}{.27} & \meansd{85.77}{.68}
        & \meansd{84.69}{.36} & \meansd{91.71}{.35} & \meansd{95.90}{.22} & \meansd{97.97}{.25} & \meansd{89.78}{.73} \\

        MALUNet \cite{malunet}
        & \meansd{80.37}{.34} & \meansd{89.11}{.33} & \meansd{95.93}{.19} & \meansd{97.70}{.24} & \meansd{88.33}{.69}
        & \meansd{81.03}{.66} & \meansd{89.52}{.64} & \meansd{95.58}{.18} & \meansd{97.35}{.26} & \meansd{89.00}{.76}
        & \meansd{86.18}{.41} & \meansd{92.58}{.39} & \meansd{96.24}{.21} & \meansd{97.44}{.23} & \meansd{92.69}{.71} \\

        EGE-UNet \cite{egeunet}
        & \meansd{83.08}{.55} & \meansd{90.76}{.52} & \meansd{96.55}{.16} & \meansd{98.22}{.21} & \meansd{89.40}{.62}
        & \meansd{79.82}{.38} & \meansd{88.78}{.40} & \meansd{95.19}{.17} & \meansd{96.48}{.23} & \meansd{90.35}{.63}
        & \meansd{87.78}{.47} & \meansd{93.48}{.44} & \meansd{96.61}{.20} & \meansd{96.81}{.24} & \meansd{96.02}{.76} \\

        VM-UNet \cite{vmunet}
        & \meansd{82.55}{.43} & \meansd{90.44}{.45} & \meansd{96.46}{.18} & \meansd{98.32}{.22} & \meansd{88.50}{.64}
        & \meansd{80.96}{.63} & \meansd{89.48}{.61} & \meansd{95.61}{.19} & \meansd{97.55}{.25} & \meansd{88.36}{.73}
        & \meansd{86.68}{.43} & \meansd{92.87}{.41} & \meansd{96.34}{.19} & \meansd{97.08}{.22} & \meansd{94.15}{.61} \\

        VM-UNet2 \cite{vmunet2}
        & \meansd{82.38}{.39} & \meansd{90.34}{.41} & \meansd{96.44}{.17} & \meansd{\textbf{98.49}}{.20} & \meansd{87.68}{.60}
        & \meansd{80.65}{.59} & \meansd{89.29}{.57} & \meansd{95.45}{.20} & \meansd{97.02}{.24} & \meansd{89.59}{.78}
        & \meansd{87.79}{.48} & \meansd{93.50}{.46} & \meansd{96.64}{.20} & \meansd{97.04}{.23} & \meansd{95.46}{.65} \\

        LightM-UNet \cite{lightmunet}
        & \meansd{81.49}{.36} & \meansd{89.80}{.36} & \meansd{96.19}{.18} & \meansd{97.86}{.23} & \meansd{88.99}{.68}
        & \meansd{80.16}{.49} & \meansd{88.99}{.47} & \meansd{95.31}{.19} & \meansd{97.05}{.25} & \meansd{88.89}{.70}
        & \meansd{88.56}{.50} & \meansd{93.93}{.48} & \meansd{96.90}{.21} & \meansd{97.60}{.24} & \meansd{94.84}{.67} \\

        LB-UNet \cite{lbunet}
        & \meansd{82.40}{.51} & \meansd{90.35}{.49} & \meansd{96.43}{.19} & \meansd{98.46}{.22} & \meansd{87.80}{.63}
        & \meansd{81.22}{.67} & \meansd{89.64}{.68} & \meansd{95.60}{.21} & \meansd{97.11}{.27} & \meansd{89.96}{.83}
        & \meansd{89.33}{.52} & \meansd{94.36}{.53} & \meansd{97.10}{.22} & \meansd{97.44}{.25} & \meansd{96.07}{.93} \\

        ULVM-UNet \cite{ultralightvmnet}
        & \meansd{83.05}{.48} & \meansd{90.74}{.47} & \meansd{96.50}{.20} & \meansd{97.89}{.24} & \meansd{90.56}{.72}
        & \meansd{80.64}{.53} & \meansd{89.29}{.54} & \meansd{95.59}{.18} & \meansd{97.84}{.26} & \meansd{87.17}{.73}
        & \meansd{88.78}{.49} & \meansd{94.06}{.51} & \meansd{97.03}{.21} & \meansd{\textbf{98.43}}{.24} & \meansd{92.91}{.99} \\ 

        \textbf{Ours}
        & \meansd{\textbf{85.55}}{.28} & \meansd{\textbf{92.21}}{.16} & \meansd{\textbf{96.98}}{.21} & \meansd{98.39}{.24} & \meansd{\textbf{91.20}}{.76}
        & \meansd{\textbf{83.60}}{.40} & \meansd{\textbf{91.07}}{.24} & \meansd{\textbf{96.13}}{.22} & \meansd{97.95}{.26} & \meansd{89.69}{.71}
        & \meansd{\textbf{90.77}}{.36} & \meansd{\textbf{95.16}}{.20} & \meansd{\textbf{97.55}}{.28} & \meansd{98.33}{.23} & \meansd{95.24}{.76} \\
        \hline
    \end{tabular}
    \caption{Results on ISIC2017, ISIC2018, and HAM10000 datasets. All results are averaged over five runs. Values are reported as mean\textsuperscript{SD} (equivalent to mean$\pm$SD). All SOTA baselines are reproduced using their publicly available implementations with identical train–val–test splits for fair comparison. (\(\uparrow\)) indicates higher is better. Best results are in bold. HD95 results are provided in the Supp. Sec. 7.1.}

    \label{tab:isic_ham_full}
\end{table*}

\begin{table}[t]
    \centering
    \footnotesize
    \setlength{\tabcolsep}{1.35mm}
    \begin{tabular}{l|ccccc}
        \hline
        Model 
        & IoU$\uparrow$ & DSC$\uparrow$ & AC$\uparrow$ & SP$\uparrow$ & SE$\uparrow$ \\
        \hline
        U-Net \cite{unet}            
        & \meansd{80.33}{.50} & \meansd{89.09}{.33} & \meansd{89.79}{.22} & \meansd{97.35}{.27} & \meansd{82.40}{.66} \\
        SCR-Net \cite{scrnet}        
        & \meansd{74.10}{.41} & \meansd{85.13}{.27} & \meansd{86.63}{.24} & \meansd{97.91}{.30} & \meansd{75.61}{.74} \\
        TransFuse \cite{transfuse} 
        & \meansd{83.54}{.58} & \meansd{91.03}{.45} & \meansd{91.43}{.25} & \meansd{97.01}{.28} & \meansd{85.98}{.70} \\ 
        UTNetV2 \cite{utnetv2}
        & \meansd{75.21}{.47} & \meansd{85.85}{.30} & \meansd{87.27}{.23} & \meansd{98.53}{.33} & \meansd{76.28}{.78} \\
        ASwin U-Net \cite{attentionswinunet} 
        & \meansd{81.41}{.52} & \meansd{89.75}{.36} & \meansd{90.45}{.21} & \meansd{98.47}{.31} & \meansd{82.63}{.72} \\
        $C^2$SDG \cite{c2sdg}         
        & \meansd{82.23}{.55} & \meansd{90.25}{.39} & \meansd{90.68}{.22} & \meansd{96.30}{.26} & \meansd{85.20}{.68} \\
        UNeXt-S \cite{unext}        
        & \meansd{82.51}{.60} & \meansd{90.41}{.42} & \meansd{90.73}{.23} & \meansd{95.21}{.25} & \meansd{86.37}{.75} \\
        MALUNet \cite{malunet}       
        & \meansd{85.99}{.62} & \meansd{92.47}{.56} & \meansd{92.76}{.26} & \meansd{97.90}{.29} & \meansd{87.75}{.80} \\
        EGE-UNet \cite{egeunet}     
        & \meansd{86.97}{.65} & \meansd{93.03}{.59} & \meansd{93.31}{.27} & \meansd{98.55}{.32} & \meansd{88.20}{.84} \\
        VM-UNet \cite{vmunet}       
        & \meansd{84.73}{.57} & \meansd{91.73}{.47} & \meansd{92.01}{.24} & \meansd{96.49}{.27} & \meansd{87.63}{.76} \\
        VM-UNet2 \cite{vmunet2}
        & \meansd{85.30}{.62} & \meansd{92.07}{.50} & \meansd{92.47}{.25} & \meansd{98.65}{.34} & \meansd{86.43}{.77} \\
        LightM-UNet \cite{lightmunet}
        & \meansd{85.38}{.63} & \meansd{92.11}{.53} & \meansd{92.45}{.26} & \meansd{97.88}{.28} & \meansd{87.14}{.79} \\
        LB-UNet \cite{lbunet}        
        & \meansd{87.12}{.64} & \meansd{93.12}{.65} & \meansd{93.41}{.28} & \meansd{\textbf{98.83}}{.35} & \meansd{88.12}{.92} \\
        ULVM-UNet \cite{ultralightvmnet} 
        & \meansd{87.10}{.61} & \meansd{93.10}{.62} & \meansd{93.31}{.27} & \meansd{97.40}{.24} & \meansd{89.31}{.99} \\
        \textbf{Ours}       
        & \meansd{\textbf{88.54}}{.48} & \meansd{\textbf{93.92}}{.27} & \meansd{\textbf{94.08}}{.29} & \meansd{97.79}{.30} & \meansd{\textbf{90.45}}{.83} \\
        \hline
    \end{tabular}
        \caption{Results on PH2 dataset. All results are averaged over five runs. All SOTA baselines are reproduced using their publicly available implementations with the same train–val–test splits for consistency. Best in bold. HD95 results are in Supp. Sec.~7.1.}
\label{tab:ph2}
\end{table}

\begin{table}[t]
    \centering
    \footnotesize
    \begin{tabular}{l|ccccc}
    \hline
    \multirow{2}{*}{Model}
      & \multicolumn{5}{c}{Average Performance} \\
    \cline{2-6}
      & IoU$\uparrow$ 
      & DSC$\uparrow$ 
      & AC$\uparrow$ 
      & SP$\uparrow$ 
      & SE$\uparrow$ \\
    \hline
    U-Net \cite{unet}          
      & 79.40 & 88.48 & 93.72 & 97.12 & 86.19 \\
    SCR-Net \cite{scrnet}      
      & 79.45 & 88.49 & 93.34 & 97.31 & 86.14 \\
    TransFuse \cite{transfuse} 
      & 81.76 & 89.95 & 94.42 & 96.44 & 90.36 \\
    UTNetV2 \cite{utnetv2}     
      & 76.71 & 86.82 & 92.43 & 96.10 & 85.99 \\
    ASwin U-Net \cite{attentionswinunet} 
      & 79.09 & 88.29 & 93.68 & 96.88 & 86.74 \\
    $C^2$SDG \cite{c2sdg}      
      & 81.08 & 89.54 & 94.07 & 96.01 & 90.29 \\
    UNeXt-S \cite{unext}       
      & 82.10 & 90.16 & 94.56 & 97.36 & 87.38 \\
    MALUNet \cite{malunet}     
      & 83.39 & 90.92 & 95.13 & 97.60 & 89.44 \\
    EGE-UNet \cite{egeunet}    
      & 84.41 & 91.51 & 95.42 & 97.52 & 90.99 \\
    VM-UNet \cite{vmunet}      
      & 83.73 & 91.13 & 95.10 & 97.36 & 89.66 \\
    VM-UNet2 \cite{vmunet2}    
      & 84.03 & 91.30 & 95.25 & 97.80 & 89.79 \\
    LightM-UNet \cite{lightmunet} 
      & 83.90 & 91.21 & 95.21 & 97.60 & 89.96 \\
    LB-UNet \cite{lbunet}      
      & 85.02 & 91.87 & 95.64 & 97.96 & 90.49 \\
    ULVM-UNet \cite{ultralightvmnet} 
      & 84.89 & 91.80 & 95.61 & 97.89 & 89.99 \\
    \textbf{Ours}              
      & \textbf{87.12} & \textbf{93.09} & \textbf{96.19} & \textbf{98.12} & \textbf{91.65} \\
    
    \hline

    \end{tabular}%
    \caption{Overall average performance across ISIC2017, ISIC2018, HAM10000, and PH2. Two more recent models, H-vmunet \cite{hvmunet} and WTCM-UNet \cite{wtcmunet}, are also compared in Supp. Sec.~7.2.}
    \label{tab:avg_performance}
\end{table}

\begin{table}[t]
    \centering
    \footnotesize
    \setlength{\tabcolsep}{1.2mm}
    \begin{tabular}{c|l|cc}
    \hline
    Model Class & Model & Params (M)$\downarrow$ & GFLOPs$\downarrow$ \\
    \hline
    C & U-Net \cite{unet}             & 7.773  & 13.758 \\
    C & SCR-Net \cite{scrnet}         & 0.801  &  1.567 \\
    T & TransFuse \cite{transfuse}    & 26.270 & 11.530 \\
    T & UTNetV2 \cite{utnetv2}        & 12.800 & 15.500 \\
    T & ASwin U-Net \cite{attentionswinunet} & 46.910 & 14.181 \\
    C & $C^2$SDG \cite{c2sdg}         & 22.001 &  7.972 \\
    C & UNeXt-S \cite{unext}          & 0.302  &  0.103 \\
    C & MALUNet \cite{malunet}        & 0.175  &  0.083 \\
    C & EGE-UNet \cite{egeunet}       & 0.053  &  0.072 \\
    M & VM-UNet \cite{vmunet}         & 27.427 &  4.112 \\
    M & VM-UNet2 \cite{vmunet2}       & 22.771 &  4.400 \\
    M & LightM-UNet \cite{lightmunet} & 0.403  &  0.391 \\
    C & LB-UNet \cite{lbunet}         & \textbf{0.038} & 0.098 \\
    M & ULVM-UNet \cite{ultralightvmnet} & 0.049  & \textbf{0.060} \\
    M & \textbf{Ours}                 & 0.494  & 0.326 \\
    \hline
    \end{tabular}
    \caption{Model complexity in parameters (M) and GFLOPs (@\(256 \times 256\)). Model Class denotes the architectural family: C = CNN, T = Transformer, M = Mamba. Best results are bold.}
    \label{tab:complexity}
\end{table}

\begin{figure}[t]
    \centering
    \includegraphics[width=0.9\linewidth]{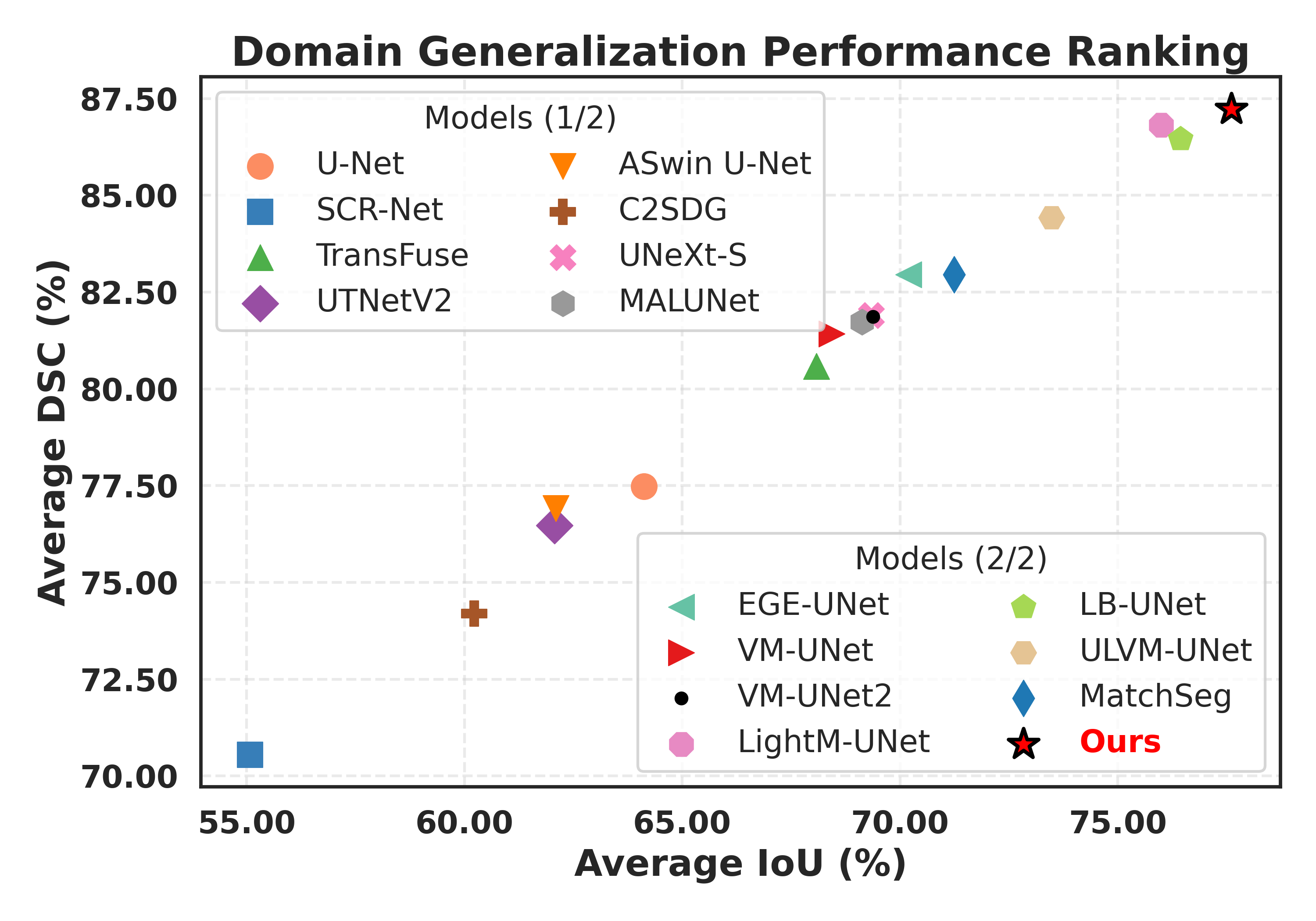}
    \caption{Domain generalization ranking (average IoU vs. average DSC) on six unseen HAM10000 lesion categories. Each marker represents a model; the best performers are in the top right corner.}
    \label{fig:cross_domain_rank}
\end{figure}

\begin{table*}[t]
\centering
\footnotesize
\setlength{\tabcolsep}{3.2pt}
\begin{tabular}{l|cc|cc|cc|cc|cc|cc|cc}
\hline
\multirow{2}{*}{Models} 
& \multicolumn{2}{c|}{AKIEC} 
& \multicolumn{2}{c|}{BCC} 
& \multicolumn{2}{c|}{BKL} 
& \multicolumn{2}{c|}{DF} 
& \multicolumn{2}{c|}{MEL} 
& \multicolumn{2}{c|}{VASC} 
& \multicolumn{2}{c}{Avg} \\
\cline{2-15}
& IoU$\uparrow$ & DSC$\uparrow$ 
& IoU$\uparrow$ & DSC$\uparrow$ 
& IoU$\uparrow$ & DSC$\uparrow$ 
& IoU$\uparrow$ & DSC$\uparrow$ 
& IoU$\uparrow$ & DSC$\uparrow$ 
& IoU$\uparrow$ & DSC$\uparrow$ 
& IoU$\uparrow$ & DSC$\uparrow$ \\
\hline
U-Net \cite{unet}              & 55.08 & 71.04 & 51.49 & 67.98 & 68.50 & 81.31 & 69.27 & 81.85 & 76.02 & 86.38 & 64.39 & 78.34 & 64.12 & 77.48 \\
SCR-Net \cite{scrnet}          & 62.48 & 76.91 & 47.81 & 64.69 & 61.84 & 76.42 & 52.29 & 68.67 & 66.43 & 79.83 & 39.64 & 56.77 & 55.08 & 70.55 \\
TransFuse \cite{transfuse}     & 70.06 & 82.39 & 58.17 & 73.55 & 76.84 & 86.90 & 64.54 & 78.45 & 83.79 & 91.18 & 55.10 & 71.05 & 68.08 & 80.59 \\
UTNetV2 \cite{utnetv2}         & 64.14 & 78.16 & 53.58 & 69.78 & 70.67 & 82.81 & 54.82 & 70.82 & 78.42 & 87.91 & 50.78 & 67.36 & 62.07 & 76.47 \\
ASwin U-Net \cite{attentionswinunet} & 53.27 & 69.51 & 48.36 & 65.19 & 65.15 & 78.90 & 64.84 & 78.67 & 75.42 & 85.99 & 65.54 & 79.18 & 62.10 & 76.91 \\
$C^2$SDG \cite{c2sdg}          & 64.49 & 78.41 & 52.33 & 68.70 & 68.57 & 81.35 & 58.89 & 74.13 & 75.91 & 86.31 & 43.15 & 60.29 & 60.22 & 74.20 \\
UNeXt-S \cite{unext}           & 63.19 & 77.45 & 56.76 & 72.41 & 76.79 & 86.87 & 70.50 & 82.70 & 83.54 & 91.03 & 65.24 & 78.96 & 69.34 & 81.90 \\
MALUNet \cite{malunet}         & 69.28 & 81.85 & 59.52 & 74.62 & 76.71 & 86.82 & 68.03 & 80.97 & 81.63 & 89.89 & 61.61 & 76.24 & 69.13 & 81.73 \\
EGE-UNet \cite{egeunet}        & 68.74 & 81.48 & 57.96 & 73.38 & 80.11 & 88.96 & 69.21 & 81.81 & 85.16 & 91.99 & 64.01 & 78.06 & 70.20 & 82.95 \\
VM-UNet \cite{vmunet}          & 70.54 & 82.73 & 59.71 & 74.77 & 77.95 & 87.61 & 64.01 & 78.05 & 84.18 & 91.41 & 56.19 & 71.95 & 68.43 & 81.42 \\
VM-UNet2 \cite{vmunet2}        & 70.42 & 82.64 & 58.48 & 73.80 & 76.41 & 86.63 & 68.84 & 81.54 & 83.04 & 90.74 & 61.09 & 75.85 & 69.38 & 81.87 \\
LightM-UNet \cite{lightmunet}  & 71.30 & 83.24 & 64.30 & 78.27 & 82.63 & 90.49 & \textbf{77.35} & \textbf{87.23} & 87.45 & 93.31 & 72.96 & 84.37 & 76.00 & 86.82 \\
LB-UNet \cite{lbunet}          & 70.99 & 83.04 & 65.38 & 79.06 & 82.43 & 90.37 & 77.12 & 87.08 & 87.48 & 93.32 & \textbf{75.21} & \textbf{85.85} & 76.44 & 86.45 \\
ULVM-UNet \cite{ultralightvmnet} & 69.39 & 81.93 & 60.58 & 75.45 & 81.50 & 89.81 & 74.15 & 85.16 & 86.84 & 92.96 & 68.44 & 81.26 & 73.48 & 84.43 \\
MatchSeg \cite{matchseg}   & 67.81 & 80.82 & 58.14 & 73.53 & 76.12 & 86.44 & 70.90 & 82.97 & 84.58 & 91.64 & 69.89 & 82.28 & 71.24 & 82.95\\
\textbf{Ours} & \textbf{74.53} & \textbf{85.41} & \textbf{67.33} & \textbf{80.47} & \textbf{83.70} & \textbf{91.13} & 76.52 & 86.70 & \textbf{88.50} & \textbf{93.90} & 75.08 & 85.77 & \textbf{77.61} & \textbf{87.23} \\
\hline
\end{tabular}
\caption{Domain generalization results on HAM10000. We train all models (except MatchSeg) for the NV lesion type only, using the same train-val-test splits, and then tested on the other six unseen categories (AKIEC, BCC, BKL, DF, MEL, and VASC) of the HAM10000 dataset. For each lesion category, we report IoU and DSC. We get MatchSeg result from \cite{matchseg}. All results are averaged over five independent runs (except MatchSeg), and the best results are highlighted in bold.}
\label{tab:cross_domain_results_nested}
\end{table*}

\subsection{Comparison with SOTA Methods}
To validate the effectiveness of MambaLiteUNet, we compare it with recent models on ISIC2017, ISIC2018, HAM10000, and PH2. Results are summarized in Table~\ref{tab:isic_ham_full} and Table~\ref{tab:ph2}, with overall averages in Table~\ref{tab:avg_performance}.

On ISIC2017, our model achieves an IoU of 85.55\% and a DSC of 92.21\%, improving over the next best ULVM-UNet by 2.50 and 1.47 points, respectively. We also lead in accuracy (96.98\%) and sensitivity (91.20\%), while maintaining competitive specificity (98.39\%). For ISIC2018, MambaLiteUNet achieves an IoU of 83.60\% and Dice of 91.07\%, outperforming ULVM-UNet by 2.96 and 1.78 points in IoU and Dice, respectively. We also rank first in accuracy (96.13\%), with specificity (97.95\%) closely trailing UNeXt-S’s 98.18\% (Table~\ref{tab:isic_ham_full}).

On the more challenging HAM10000, our model similarly excels, achieving an IoU of 90.77\% and a DSC of 95.16\%, outperforming LB-UNet by 1.44 and 0.80 points and ULVM-UNet by 1.99 and 1.10 points, respectively. It achieves the highest accuracy (97.55\%), while ranking second in specificity (98.33\%), just behind ULVM-UNet (98.43\%) (Table~\ref{tab:isic_ham_full}). On PH2, we achieve the highest IoU (88.54\%) and DSC (93.92\%), with improvements of 1.42 and 0.80 points over LB-UNet, while also leading in accuracy (94.08\%) and sensitivity (90.45\%) (Table~\ref{tab:ph2}).

Across all four datasets, MambaLiteUNet achieves the best average performance, surpassing LB-UNet by 2.10 points in IoU and 1.22 points in DSC (Table~\ref{tab:avg_performance}). It also ranks best on AC, SP, and SE. Our model achieves this performance with 0.494M parameters and 0.326 GFLOPs, showing a strong balance between segmentation quality and efficiency (Table~\ref{tab:complexity}).

\begin{figure*}[t]
    \centering
    \includegraphics[width=0.83\textwidth]{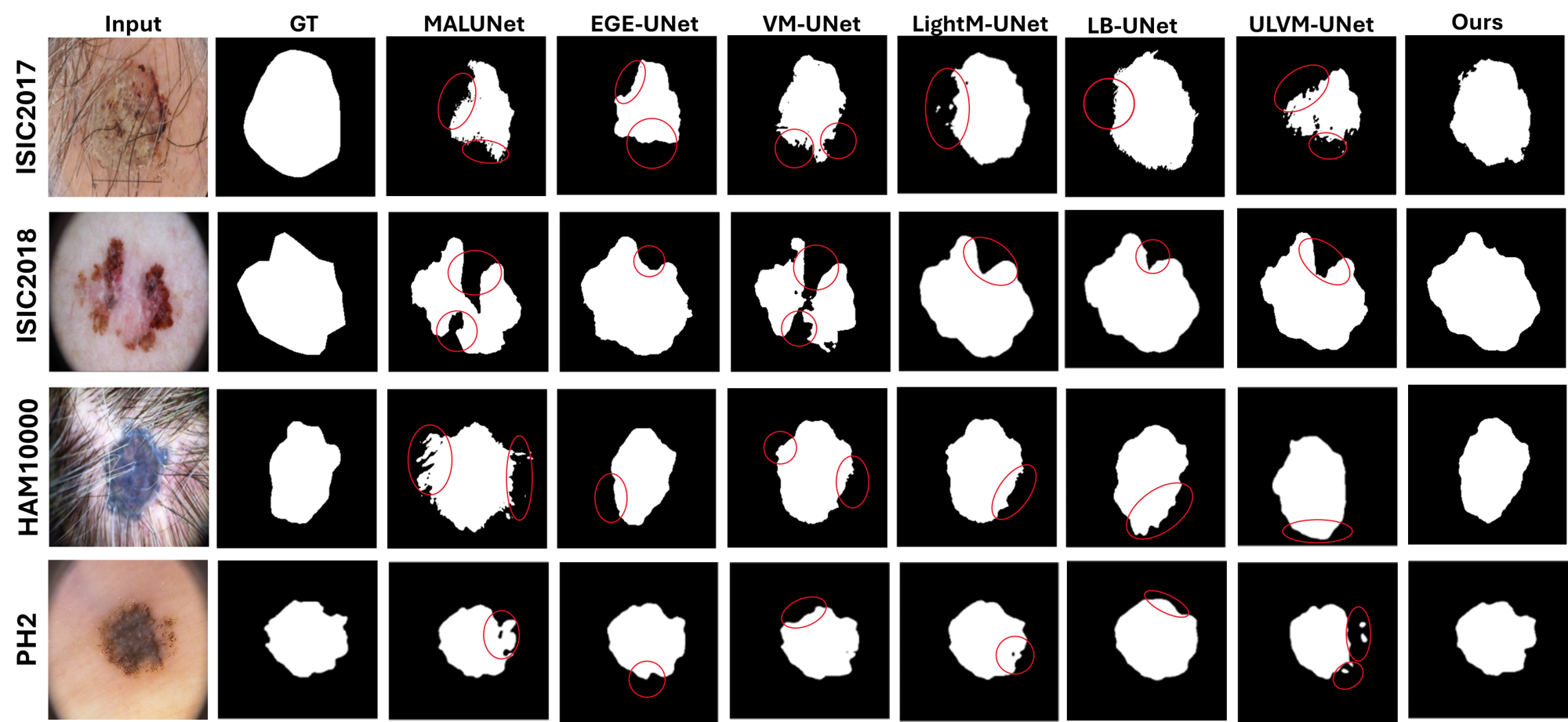}
    \caption{Qualitative comparison on ISIC2017, ISIC2018, HAM10000, and PH2. Red outlines highlight segmentation errors and weaknesses of SOTA methods, demonstrating our method’s superiority.}
    \label{fig:qualitative_comparison}
\end{figure*}

A qualitative comparison in Figure~\ref{fig:qualitative_comparison} further shows that our method captures fine boundary details that existing models miss, particularly in regions of low contrast, hair occlusions, and irregular shapes. Finally, the complexity--performance trade-off plot in Figure~\ref{fig:tradeoff_plot} shows that MambaLiteUNet lies in the upper-left region, indicating both high accuracy and low complexity, thereby highlighting its practical utility for resource-constrained applications.

\subsection{Domain Generalization Ability}
We evaluate the generalization ability of MambaLiteUNet under distribution shift on HAM10000. Following~\cite{matchseg}, all models are trained only on the nevus (NV) lesion type. NV is used as the source domain because it has weak boundary contrast, making it a challenging class for learning transferable representations. The trained models are then evaluated on six unseen categories: AKIEC, BCC, BKL, DF, MEL, and VASC. This protocol reflects a realistic clinical setting where only a limited set of lesion types is available during training, while reliable segmentation is required across diverse patterns at test time.
Table~\ref{tab:cross_domain_results_nested} shows that MambaLiteUNet achieves the best average performance, reaching 77.61\% IoU and 87.23\% DSC, and outperforming all SOTA methods on four of the six categories. In particular, the model achieves 93.90\% DSC on MEL and 91.13\% on BKL, both of which are clinically challenging due to highly irregular lesion boundaries. On the remaining two categories, the performance is also close to the best results. These results indicate that our model maintains strong robustness under domain shifts across diverse lesion types. Figure~\ref{fig:cross_domain_rank} shows the average IoU--DSC ranking under this domain-generalization setting.

We also evaluate cross-dataset generalization from ISIC2018 to PH2, test generalization beyond dermoscopic images on BUS (ultrasound)~\cite{bus} and GlaS (histopathology)~\cite{glas}, and analyze robustness under limited training data. These results are provided in Supp. Sec.~7.3--7.5.

\begin{table*}[t]
\centering
\renewcommand{\arraystretch}{1.0}
\resizebox{0.93\textwidth}{!}{%
\begin{tabular}{cc|ccccc|ccccc|ccccc}
\hline
\multicolumn{2}{c|}{Loss} 
& \multicolumn{5}{c|}{ISIC2017} 
& \multicolumn{5}{c|}{ISIC2018} 
& \multicolumn{5}{c}{HAM10000} \\
\hline
BCE & Dice
& IoU$\uparrow$ & DSC$\uparrow$ & AC$\uparrow$ & SP$\uparrow$ & SE$\uparrow$
& IoU$\uparrow$ & DSC$\uparrow$ & AC$\uparrow$ & SP$\uparrow$ & SE$\uparrow$
& IoU$\uparrow$ & DSC$\uparrow$ & AC$\uparrow$ & SP$\uparrow$ & SE$\uparrow$ \\
\hline
\checkmark &           & 84.88 & 91.82 & 96.87 & \textbf{98.66} & 89.54 
                       & 82.51 & 90.42 & 95.85 & 97.72 & 89.20
                       & 90.23 & 94.87 & 97.41 & \textbf{98.32} & 94.71 \\
          & \checkmark & 85.15 & 91.98 & 96.91 & 98.48 & \textbf{90.48} 
                       & 81.12 & 89.58 & 95.38 & 96.81 & \textbf{90.30}
                       & 90.55 & 95.04 & 97.48 & 98.15 & \textbf{95.51} \\
\checkmark & \checkmark & \textbf{85.55} & \textbf{92.21} & \textbf{96.98} & 98.39 & \textbf{91.20} 
                       & \textbf{83.60} & \textbf{91.07} & \textbf{96.13} & \textbf{97.95} & 89.69
                       & \textbf{90.77} & \textbf{95.16} & \textbf{97.55} & 98.33 & 95.24 \\
\hline
\end{tabular}%
}
\caption{Performance of different losses on ISIC2017, ISIC2018, and HAM10000. \checkmark\ indicates the loss selection. Best results are in bold.}
\label{tab:loss_ablation}
\end{table*}

\begin{table*}[t]
\centering
\footnotesize

\setlength{\tabcolsep}{4pt}
\renewcommand{\arraystretch}{1.0}
\begin{tabular}{ccc|cc|ccccc|ccccc}
\hline
\multicolumn{3}{c|}{Modules with Mamba} 
 & \multicolumn{2}{c|}{Complexity} 
 & \multicolumn{5}{c|}{ISIC2017} 
 & \multicolumn{5}{c}{ISIC2018} \\
\cline{1-15}
AMF & LGFM & CGA 
 & Params (M)$\downarrow$ & GFLOPs$\downarrow$ 
 & IoU$\uparrow$ & DSC$\uparrow$ & AC$\uparrow$ & SP$\uparrow$ & SE$\uparrow$
 & IoU$\uparrow$ & DSC$\uparrow$ & AC$\uparrow$ & SP$\uparrow$ & SE$\uparrow$ \\

\hline
           &           &           & 0.425 & 0.938 & 82.45 & 90.38 & 96.49 & 98.68 & 87.12 & 80.59 & 89.25 & 95.35 & 97.44 & 87.91 \\
\checkmark &           &           & 0.226 & 0.830 & 84.35 & 91.51 & 96.74 & 98.42 & 89.81 & 82.57 & 90.45 & 95.86 & 97.74 & 89.20 \\
           & \checkmark &           & 0.180 & 0.794 & 84.88 & 91.82 & 96.85 & 98.45 & 90.27 & 82.25 & 90.26 & 95.70 & 97.10 & \textbf{90.73} \\
           &           & \checkmark & 0.593 & 0.478 & 84.68 & 91.71 & 96.77 & 98.15 & 91.11 & 82.61 & 90.48 & 95.87 & 97.75 & 89.22 \\
\checkmark & \checkmark &           & 0.326 & 0.238 & 85.23 & 92.03 & 96.97 & \textbf{98.81} & 89.39 & 82.90 & 90.65 & 95.92 & 97.58 & 90.02 \\
\checkmark &           & \checkmark & 0.395 & 0.311 & 85.22 & 92.02 & 96.94 & 98.58 & 90.21 & 83.28 & 90.87 & 96.04 & 97.83 & 89.70 \\
           & \checkmark & \checkmark & 0.350 & 0.305 & 85.21 & 92.02 & 96.94 & 98.66 & 89.91 & 83.07 & 90.75 & 95.95 & 97.47 & 90.52 \\
\checkmark & \checkmark & \checkmark & 0.494 & 0.326 & \textbf{85.55} & \textbf{92.21} & \textbf{96.98} & 98.39 & \textbf{91.20} & \textbf{83.60} & \textbf{91.07} & \textbf{96.13} & \textbf{97.95} & 89.69 \\
\hline
\end{tabular}%
\caption{Ablation of AMF, LGFM, and CGA on ISIC2017 and ISIC2018. 
A Mamba-free control study is provided in Supp. Sec.~8.2.}
\label{tab:module_contributions1}
\end{table*}

\begin{table}[t]
\centering
\footnotesize
\renewcommand{\arraystretch}{1.1}
\setlength{\tabcolsep}{3pt}
\begin{tabular}{c|cc|ccccc}
\hline
Branch 
  & Params (M)$\downarrow$ & GFLOPs$\downarrow$ 
  & IoU$\uparrow$ & DSC$\uparrow$ & AC$\uparrow$ & SP$\uparrow$ & SE$\uparrow$ \\
\hline
1  & \textbf{0.381} & \textbf{0.270} & 81.74 & 89.95 & 95.60 & 97.29 & \textbf{89.62} \\
2  & 0.418          & 0.289          & 82.50 & 90.41 & 95.82 & 97.57 & \textbf{89.62} \\
4  & 0.494          & 0.326          & \textbf{83.60} & \textbf{91.07} & \textbf{96.13} & \textbf{97.95} & 89.69 \\
8  & 0.646          & 0.402          & 82.38 & 90.34 & 95.79 & 97.55 & 89.57 \\
16 & 0.949          & 0.553          & 81.24 & 89.65 & 95.53 & 97.64 & 88.04 \\
\hline
\end{tabular}
\caption{Results for different branch configurations on ISIC2018.}
\label{tab:branch_config}
\end{table}

\begin{table}[!h]
\centering
\footnotesize
\renewcommand{\arraystretch}{1.0}
\setlength{\tabcolsep}{3pt}
\begin{tabular}{l|cc|ccccc}
\hline
CC & Params (M)$\downarrow$ & GFLOPs$\downarrow$ 
   & IoU$\uparrow$ & DSC$\uparrow$ & AC$\uparrow$ & SP$\uparrow$ & SE$\uparrow$ \\
\hline
C1 & \textbf{0.131} & \textbf{0.092} & 82.10 & 90.17 & 95.70 & 97.41 & 89.66 \\
C2 & 0.242          & 0.118          & 82.88 & 90.64 & 95.91 & 97.57 & \textbf{90.01} \\
C3 & 0.494          & 0.326          & \textbf{83.60} & \textbf{91.07} & \textbf{96.13} & \textbf{97.95} & 89.69 \\
C4 & 0.878          & 0.168          & 81.31 & 89.69 & 95.48 & 97.19 & 89.41 \\
C5 & 3.439          & 0.622          & 82.52 & 90.42 & 95.87 & 97.88 & 88.72 \\
\hline
\end{tabular}
\caption{Different Channel Configurations (CC) on ISIC2018.}
\label{tab:channels_input_size_config}
\end{table}

\begin{table}[!h]
\centering
\footnotesize
\renewcommand{\arraystretch}{1.0}
\setlength{\tabcolsep}{3pt}
\resizebox{\linewidth}{!}{%
\begin{tabular}{l|cc|ccccc}
\hline
Image Size 
   & Params (M)$\downarrow$ & GFLOPs$\downarrow$ 
   & IoU$\uparrow$ & DSC$\uparrow$ & SE$\uparrow$ & SP$\uparrow$ & AC$\uparrow$ \\
\hline
$224\times224$ & \textbf{0.494} & \textbf{0.250} & 83.08 & 90.76 & 96.01 & 97.91 & 89.24 \\
$256\times256$ & 0.494          & 0.326          & \textbf{83.60} & \textbf{91.07} & \textbf{96.13} & 97.95 & \textbf{89.69} \\
$288\times288$ & 0.494          & 0.413          & 81.93 & 90.07 & 95.69 & 97.60 & 88.92 \\
$320\times320$ & 0.494          & 0.510          & 82.42 & 90.36 & 95.84 & 97.86 & 88.69 \\
$512\times512$ & 0.494          & 1.305          & 83.33 & 90.91 & 96.10 & \textbf{98.17} & 88.75 \\
\hline
\end{tabular}%
}
\caption{Impact of input size on ISIC2018 performances.}
\label{tab:input_size_config}
\end{table}

\subsection{Ablation Study}
In this section, we perform ablation studies to evaluate the impact of MambaLiteUNet’s key design choices.

\subsubsection{Effect of Different Loss Functions}
We perform an ablation study to assess BCE loss, Dice loss, and their combination on ISIC2017, ISIC2018, HAM10000, and PH2 (see Supp. Sec.~8.1). As shown in Table \ref{tab:loss_ablation}, the combined loss consistently outperforms both individual losses across all datasets in terms of IoU and DSC. These results indicate that combining pixel-wise and region-based supervision improves the model’s ability to learn precise boundaries and complete lesion structures.

\subsubsection{Effect of Different Core Architectural Modules}
We evaluate each module's impact on ISIC2017 and ISIC2018 (Table~\ref{tab:module_contributions1}). The baseline (no modules) achieves 82.45\% IoU / 90.38\% DSC on ISIC2017. Adding AMF alone reduces parameters by 47\% and raises IoU by 1.90 points and DSC by 1.13 points. LGFM alone boosts IoU by 2.43 points and sensitivity by 3.15 points. CGA alone achieves a 2.23 points IoU and 1.33 points DSC gain. Combining all three modules gains the best tradeoff: 85.55\% IoU / 92.21\% DSC on ISIC2017 and 83.60\% IoU / 91.07\% DSC on ISIC2018, using only 0.494M parameters and 0.326 GFLOPs. These results show that AMF, LGFM, and CGA work together to optimize segmentation accuracy while maintaining low computational costs.

\subsubsection{Effect of Different Number of Branches}
We study the effect of varying the number of parallel branches in the AMF and CGA modules from 1 to 16. As shown in Table~\ref{tab:branch_config}, performance improves up to four branches, then declines. Four branches achieve the highest DSC of 91.07\% and IoU of 83.60\% on ISIC2018 in both modules. Fewer branches limit representational diversity, whereas more branches increase redundancy, overfitting, and complexity. Overall, four branches offer the best performance--efficiency trade-off.

\subsubsection{Effect of Different Channels and Input Sizes}
We analyze how varying the channels and input resolution affects the overall performance of the model on ISIC2018. 
We examine channel configurations (C1: $\{8, 16, 24, 32, 48, 64\}$, C2: $\{8, 16, 32, 48, 64, 96\}$, C3: $\{16, 32, 48, 64, 96, 128\}$, C4: $\{8, 16, 32, 64, 128, 256\}$, C5: $\{16, 32, 64, 128, 256, 512\}$), ranging from lightweight (C1: 0.131M params) to heavy (C5: 3.439M params). As shown in Table~\ref{tab:channels_input_size_config}, performance improves with increased capacity up to configuration C3 (\{16, 32, 48, 64, 96, 128\}), which achieves the best IoU and DSC scores. Beyond this point, larger configurations (C4 and C5) add substantial computational overhead without reliable benefits.

We also evaluate input sizes from $224 \times 224$ to $512 \times 512$, shown in Table \ref{tab:input_size_config}. While larger input slightly improves specificity, they require significantly higher GFLOPs. The $256 \times 256$ size offers the best DSC (91.07\%) with balanced complexity, making it the most efficient setting. Therefore, carefully selecting channels and input sizes is crucial for optimizing performance within computational limits.
\section{Conclusion}
In this paper, we have introduced MambaLiteUNet, a robust segmentation framework designed to learn fine lesion details within strict computational constraints. MambaLiteUNet employs AMF for rich multi-scale representation, LGFM for precise texture-to-context integration, and CGA for selective skip connection refinement. Extensive evaluation demonstrates our model achieves 92.21\% Dice and 85.55\% IoU on ISIC2017, 91.07\% Dice and 83.60\% IoU on ISIC2018, 95.16\% Dice and 90.77\% IoU on HAM10000, and 93.92\% Dice and 88.54\% IoU on PH2, with an average Dice of 93.09\% and average IoU of 87.12\%. It outperforms SOTA models by a large margin. In domain generalization, MambaLiteUNet also scores an average of 87.23\% Dice and 77.61\% IoU on six unseen lesion types, outperforming all compared models and demonstrating strong generalization ability. We believe MambaLiteUNet can be a valuable tool for improving various medical and semantic segmentation tasks. We plan to extend this work to 3D volumetric segmentation for CT and MRI analysis.

{
    \small

}


\clearpage
\setcounter{page}{1}
\maketitlesupplementary
\renewcommand{\meansd}[2]{#1\textsuperscript{#2}}

\section{Evaluation Metrics}
We evaluate our segmentation performance using overlap-based, boundary-based, and classification-based metrics. The Intersection over Union (IoU), also known as the Jaccard index, calculates the ratio of the intersection between the predicted and ground truth masks relative to their union. The Dice similarity coefficient (DSC), which is equivalent to the F1 score, emphasizes correct overlaps by giving twice the weight to true positives. For boundary quality, we use the 95th percentile Hausdorff Distance (HD95), which measures the alignment of lesion contours while reducing the influence of outliers. Additionally, we report Accuracy (AC), Sensitivity (SE), and Specificity (SP) to reflect pixel-level classification. The metrics are defined as follows:

\begin{align}
\mathrm{IoU}   &= \frac{TP}{TP + FP + FN},             \\
\mathrm{DSC}   &= \frac{2\,TP}{2\,TP + FP + FN},       \\
\mathrm{AC}   &= \frac{TP + TN}{TP + TN + FP + FN},   \\
\mathrm{SE}    &= \frac{TP}{TP + FN},                  \\
\mathrm{SP}    &= \frac{TN}{TN + FP},                  \\
\mathrm{HD}(G, P) &=  \max(
\max_{g \in G} \min_{\hat{p} \in P} d(g, \hat{p}),
\max_{\hat{p} \in P} \min_{g \in G} d(\hat{p}, g))
\end{align}
where $TP$, $FP$, $FN$, and $TN$ represent true positives, false positives, false negatives, and true negatives, respectively.
$G$ and $P$ denote the ground-truth and predicted masks, and $d(\cdot,\cdot)$ is the Euclidean distance between boundary points. HD95 corresponds to the 95th percentile of these distances, providing a robust measure of boundary alignment.

\section{Additional Experiments and Results} 
\label{supp:add_exp_results} 

To complement the main paper, we present additional experiments that further broaden the evaluation and justification of our framework. These cover boundary-focused metrics such as HD95, cross-dataset generalization, and tests on non-dermoscopic datasets.

\subsection{HD95 Evaluation across Four Datasets}
To evaluate boundary accuracy, we present HD95 results on ISIC2017 \cite{isic2017challenge}, ISIC2018 \cite{isic2018challenge}, HAM10000 \cite{ham10000}, and PH2 \cite{ph2dataset}, as shown in Table~\ref{tab:hd95_four_datasets}. The HD95 measures the boundary accuracy, which is crucial in medical image segmentation, where precise lesion contours are as important as region overlap.

Traditional CNN-based models such as U-Net and SCR-Net exhibit significant errors, indicating poor boundary localization. Transformer-based methods, including TransFuse, UTNetV2, and ASwin U-Net, reduce errors in some cases but still face limitations in accuracy. More compact CNN–MLP hybrid models, such as MALUNet, EGE-UNet, LB-UNet, and ULVM-UNet, demonstrate better performance, with LB-UNet achieving 9.72 on HAM10000 and ULVM-UNet achieving 12.40 on PH2. Recent Mamba-based architectures (VM-UNet, VM-UNet2, LightM-UNet, ULVM-UNet) show clear improvements in boundary detection over CNN and Transformer models, reducing HD95 to the 12–15-pixel range, yet they still struggle to achieve consistent accuracy across datasets.

Our model achieves the lowest HD95 scores on all four datasets: 10.73 on ISIC2017, 12.94 on ISIC2018, 8.65 on HAM10000, and 9.88 on PH2. These results reduce boundary error by approximately 1.1--2.5 pixels compared to the best-performing baselines, including recent Mamba-based models. The consistent gains demonstrate that our framework produces sharper lesion boundaries across all four benchmarks, ensuring reliable segmentation performance.

\begin{table}[t]
\centering
\footnotesize
\setlength{\tabcolsep}{1.5mm}
\begin{tabular}{l|cccc}
\hline
\multirow{2}{*}{Model} & \multicolumn{4}{c}{HD95 ($\downarrow$)} \\
\cline{2-5}
& ISIC2017 & ISIC2018 & HAM10000 & PH2 \\
\hline
U-Net \cite{unet}                 & \meansd{16.48}{1.50} & \meansd{19.67}{1.65} & \meansd{15.35}{1.08} & \meansd{18.40}{1.60} \\
SCR-Net \cite{scrnet}             & \meansd{17.06}{1.52} & \meansd{15.82}{1.32} & \meansd{13.90}{0.98} & \meansd{22.80}{1.80} \\
TransFuse \cite{transfuse}               & \meansd{15.04}{1.32} & \meansd{16.77}{1.42} & \meansd{14.76}{1.00} & \meansd{16.10}{1.28} \\
UTNetV2 \cite{utnetv2}                   & \meansd{17.22}{1.55} & \meansd{17.23}{1.50} & \meansd{18.50}{1.22} & \meansd{22.00}{1.75} \\
ASwin U-Net \cite{attentionswinunet} & \meansd{15.84}{1.48} & \meansd{19.79}{1.68} & \meansd{16.17}{1.15} & \meansd{18.10}{1.55} \\
$C^2$SDG \cite{c2sdg}                    & \meansd{14.30}{1.25} & \meansd{15.21}{1.28} & \meansd{16.00}{1.08} & \meansd{17.30}{1.42} \\
UNeXt-S \cite{unext}              & \meansd{14.30}{1.18} & \meansd{15.03}{1.20} & \meansd{14.20}{0.95} & \meansd{16.85}{1.28} \\
MALUNet \cite{malunet}                   & \meansd{14.66}{1.12} & \meansd{14.72}{1.12} & \meansd{13.70}{0.92} & \meansd{15.10}{1.10} \\
EGE-UNet \cite{egeunet}           & \meansd{12.49}{1.02} & \meansd{15.40}{1.35} & \meansd{12.97}{0.86} & \meansd{14.90}{1.02} \\
VM-UNet \cite{vmunet}             & \meansd{14.43}{1.08} & \meansd{14.31}{1.18} & \meansd{13.40}{0.90} & \meansd{15.90}{1.20} \\
VM-UNet2 \cite{vmunet2}           & \meansd{14.06}{1.08} & \meansd{14.77}{1.20} & \meansd{12.84}{0.86} & \meansd{15.30}{1.15} \\
LightM-UNet \cite{lightmunet}     & \meansd{13.80}{1.12} & \meansd{15.10}{1.24} & \meansd{12.52}{0.80} & \meansd{15.40}{1.12} \\
LB-UNet \cite{lbunet}             & \meansd{12.05}{1.02} & \meansd{14.61}{1.10} & \meansd{ 9.72}{0.74} & \meansd{14.70}{1.00} \\
ULVM-UNet \cite{ultralightvmnet}  & \meansd{12.93}{0.98} & \meansd{15.06}{1.20} & \meansd{12.23}{0.88} & \meansd{12.40}{1.05} \\
\textbf{Ours}                            & \meansd{\textbf{10.73}}{0.92} & \meansd{\textbf{12.94}}{1.02} & \meansd{\textbf{8.65}}{0.62} & \meansd{\textbf{9.88}}{0.90} \\
\hline
\end{tabular}
\caption{HD95 (in pixels) results on ISIC2017, ISIC2018, HAM10000, and PH2. All results are averaged over five runs. Values are reported as mean\textsuperscript{SD} (equivalent to mean$\pm$SD). All SOTA baselines are reproduced using their publicly available implementations with identical train–val–test splits for fair comparison. (\(\downarrow\)) indicates Lower is better. Best results are in bold.}
\label{tab:hd95_four_datasets}
\end{table}

\subsection{Additional Comparison with Recent Mamba-based Segmentation Models} \label{sec:recent_mamba_models}
Table~\ref{tab:recent_mamba_models} provides an additional comparison with two recent Mamba-Based segmentation models, H-vmunet \cite{hvmunet}, and WTCM-UNet \cite{wtcmunet}, evaluated under the same protocol. Although both methods use substantially larger model capacity, MambaLiteUNet remains more effective and efficient. On average, our model surpasses H-vmunet by 1.41 points in IoU and 0.81 points in DSC while reducing HD95 by 1.20 pixels. Compared with WTCM-UNet, our model improves IoU and DSC by 3.58 and 2.09 points, respectively, and reduces HD95 by 4.06 pixels, while using much fewer parameters and GFLOPs.

\begin{table*}[t]
\centering
\footnotesize
\setlength{\tabcolsep}{3pt}
\renewcommand{\arraystretch}{1.1}
\resizebox{\textwidth}{!}{%
\begin{tabular}{l|cc|ccc|ccc|ccc|ccc|c|c}
\hline
\multirow{2}{*}{Model} & \multirow{2}{*}{P(M)$\downarrow$} & \multirow{2}{*}{F(G)$\downarrow$}
& \multicolumn{3}{c|}{ISIC2017}
& \multicolumn{3}{c|}{ISIC2018}
& \multicolumn{3}{c|}{HAM10000}
& \multicolumn{3}{c|}{PH2}
& \multirow{2}{*}{\makecell{Ours$-$Model(Avg.)\\IoU/DSC/HD95}}
& \multirow{2}{*}{\makecell{Cost vs Ours\\Params$\times$/GFLOPs$\times$}} \\
\cline{4-16}
& & 
& IoU$\uparrow$ & DSC$\uparrow$ & HD95$\downarrow$
& IoU$\uparrow$ & DSC$\uparrow$ & HD95$\downarrow$
& IoU$\uparrow$ & DSC$\uparrow$ & HD95$\downarrow$
& IoU$\uparrow$ & DSC$\uparrow$ & HD95$\downarrow$
& & \\
\hline
H-vmunet \cite{hvmunet}
& 8.97
& 0.742
& 84.22 & 91.43 & 12.81
& 81.78 & 89.98 & 14.67
& 89.54 & 94.48 & 9.47
& 87.30 & 93.22 & 10.06
& +1.41 / +0.81 / -1.20
& 18.2$\times$ / 2.3$\times$ \\
WTCM-UNet \cite{wtcmunet}
& 28.74
& 3.12
& 80.21 & 89.02 & 15.67
& 80.90 & 89.44 & 15.24
& 86.31 & 92.65 & 12.75
& 86.72 & 92.89 & 14.76
& +3.58 / +2.09 / -4.06
& 58.2$\times$ / 9.6$\times$ \\
\hline
\end{tabular}%
}
\caption{Comparison with recent Mamba-based segmentation models. H-vmunet and WTCM-UNet are re-implemented and evaluated under our unified training and evaluation protocol for fair comparison. IoU/DSC are reported in \%, and HD95 is reported in pixels. Params (M) and GFLOPs are denoted by P(M) and F(G), respectively.}
\label{tab:recent_mamba_models}
\end{table*}

\subsection{Cross-Dataset Generalization Analysis}

We evaluate cross-dataset generalization by training on ISIC2018 (train split) and directly testing on PH2 (whole dataset) without fine-tuning. This setting assesses whether models can reliably segment images of the same modality collected at different centers under varying acquisition conditions, providing a rigorous measure of domain robustness relevant to real-world applications.

As shown in Table~\ref{tab:isic2ph2}, U-Net obtains 77.02\% IoU, 87.02\% DSC, and 22.95 HD95, showing limited transferability. TransFuse (80.56\% IoU, 89.23\% DSC, 18.70 HD95) and UTNetV2 (79.94\% IoU, 88.85\% DSC, and 18.82 HD95) improve overlap but remain inconsistent.  Recent methods EGE-UNet (81.11\%  IoU, 89.57\%  DSC, 17.36 HD95) and ULVM-UNet (81.35\%  IoU, 89.72\%  DSC, 17.07 HD95) narrow the gap but still suffer from boundary-precision errors.

Our model achieves the best overall results with 81.71\% IoU, 89.93\% DSC, 93.19\% accuracy, and 15.58 HD95, outperforming CNN-, Transformer-, and Mamba-based baselines. Compared with ULVM-UNet, we gain +0.36 IoU, +0.21 DSC, and a reduction of 1.49 HD95. Compared with EGE-UNet, the improvements are +0.60 IoU, +0.36 DSC, and a reduction of 1.78 in HD95. Compared with LightM-UNet, we reduce HD95 by 1.05 while maintaining IoU and DSC. These improvements highlight stronger overlap accuracy and sharper boundaries under domain shift.

\begin{table}[t]
    \centering
    \footnotesize
    \setlength{\tabcolsep}{1.5mm}
    \begin{tabular}{l|ccccc|c}
        \hline
        \multirow{2}{*}{Model} 
        & \multicolumn{6}{c}{Train on ISIC2018 $\rightarrow$ Test on PH2} \\
        \cline{2-7}
        & IoU$\uparrow$ & DSC$\uparrow$ & AC$\uparrow$ & SP$\uparrow$ & SE$\uparrow$ & HD95$\downarrow$ \\
        \hline
		U-Net \cite{unet} & 77.02 & 87.02 & 90.89 & 89.11 & 94.63 & 22.95 \\
		SCR-Net \cite{scrnet} & 78.93 & 88.23 & 92.17 & 92.76 & 90.94 & 19.54 \\
        ASwin U-Net \cite{attentionswinunet} & 75.01 & 85.72 & 90.53 & 91.71 & 88.07 & 21.76 \\
        TransFuse \cite{transfuse} & 80.56 & 89.23 & 92.58 & 91.29 & \textbf{95.29} & 18.70 \\
        UTNetV2 \cite{utnetv2} & 79.94 & 88.85 & 92.69 & \textbf{93.86} & 90.23 & 18.82 \\
		$C^2$SDG \cite{c2sdg} & 79.83 & 88.79 & 92.29 & 91.22 & 94.54 & 21.53 \\
		UNeXt-S \cite{unext} & 80.70 & 89.32 & 92.71 & 91.85 & 94.52 & 18.42 \\
		MALUNet \cite{malunet} & 79.87 & 88.81 & 92.46 & 92.32 & 92.74 & 19.62 \\
		EGE-UNet \cite{egeunet} & 81.11 & 89.57 & 93.08 & 93.56 & 92.07 & 17.36 \\
        VM-UNet \cite{vmunet} & 80.75 & 89.35 & 92.79 & 92.34 & 93.74 & 17.12 \\
        VM-UNet2 \cite{vmunet2} & 80.94 & 89.47 & 92.76 & 91.58 & 95.25 & 17.76 \\
        LightM-UNet \cite{lightmunet} & 81.10 & 89.56 & 92.97 & 92.71 & 93.52 & 16.63 \\       
        LB-UNet \cite{lbunet} & 81.17 & 89.61 & 92.91 & 92.03 & 94.76 & 17.38 \\
        ULVM-UNet \cite{ultralightvmnet} & 81.35 & 89.72 & 92.96 & 91.90 & 95.18 & 17.07 \\
		\textbf{Ours} & \textbf{81.71} & \textbf{89.93} & \textbf{93.19} & 92.68 & 94.26 & \textbf{15.58} \\
        \hline
    \end{tabular}
    \caption{Cross-dataset generalization performance when training on ISIC2018 and testing on the full PH2 dataset without fine-tuning. (\(\uparrow\)) indicates higher is better. (\(\downarrow\)) indicates lower is better. Best results are in bold.}
    \label{tab:isic2ph2}
\end{table}

\subsection{Generalization to Non-Dermoscopic Datasets}
To evaluate generalization beyond dermoscopic images, we extend our analysis to the BUS \cite{bus} and GlaS \cite{glas} datasets. BUS contains breast ultrasound scans with heavy speckle noise, low contrast, and irregular lesion boundaries, while GlaS comprises colorectal histopathology images characterized by structural complexity and staining variability. Both datasets present substantially different challenges compared to dermoscopic benchmarks.
Table~\ref{tab:bus_glas_generalization} summarizes the results. On BUS, U-Net achieves 67.03\% IoU, 80.26\% DSC, and 22.72 HD95, while EGE-UNet records 65.81\% IoU, 79.38\% DSC, and 19.29 HD95. Transformer-based methods perform better but remain inconsistent. TransFuse obtains 70.16\% IoU, 82.46\% DSC, and 18.46 HD95 with sensitivity at 79.68, while UTNetV2 achieves 68.63\% IoU, 81.40\% DSC, and 25.15 HD95 with sensitivity at 86.81. Mamba-based architectures improve boundary accuracy, with VM-UNet reaching 72.02\% IoU, 83.74\% DSC, and 15.29 HD95, and LightM-UNet 71.44\% IoU, 83.34\% DSC, and 15.37 HD95, but their overlap scores remain limited.  

Our model achieves 77.68\% IoU, 87.44\% DSC, and 11.55 HD95 on BUS, improving over the strongest baseline ($C^2$SDG: 73.11 IoU, 84.47 DSC, 13.32 HD95) by +4.57 IoU, +2.97 DSC, and a reduction of 1.77 in HD95. This performance underlines the model’s ability to retain both overlap accuracy and precise boundary localization under heavy noise and low contrast. On GlaS, MALUNet delivers 74.64\% IoU, 85.48\% DSC, and 24.17 HD95, while our model achieves 78.63\% IoU, 88.04\% DSC, and 21.62 HD95, improving by +3.99 IoU, +2.56 DSC, and a reduction of 2.55 in HD95. Here, the gains show that our approach adapts to structural irregularities and staining variations that cause other baselines, including Mamba-based ones, to degrade. 
Therefore, our model demonstrates robustness across imaging modalities by maintaining consistent improvements on BUS and GlaS.

\begin{table*}[t]
    \centering
    \footnotesize
    \begin{tabular}{l|ccccc|c||ccccc|c}
        \hline
        \multirow{2}{*}{Model} 
        & \multicolumn{6}{c||}{BUS (Ultrasound) \cite{bus}}
        & \multicolumn{6}{c}{GlaS (Histopathology) \cite{glas}} \\
        \cline{2-13}
        & IoU$\uparrow$ & DSC$\uparrow$ & AC$\uparrow$ & SP$\uparrow$ & SE$\uparrow$ & HD95$\downarrow$ 
        & IoU$\uparrow$ & DSC$\uparrow$ & AC$\uparrow$ & SP$\uparrow$ & SE$\uparrow$ & HD95$\downarrow$ \\
        \hline
        U-Net~\cite{unet}                 
        & 67.03 & 80.26 & 97.95 & 98.31 & 90.46 & 22.72
        & 72.69 & 84.19 & 83.86 & 83.75 & 83.96 & 25.30 \\
        TransFuse~\cite{transfuse}        
        & 70.16 & 82.46 & 98.44 & 99.34 & 79.68 & 18.46
        & 73.49 & 84.72 & 83.47 & 77.10 & \textbf{89.55} & 25.49 \\
        UTNetV2~\cite{utnetv2}            
        & 68.63 & 81.40 & 98.17 & 98.72 & 86.81 & 25.15
        & 67.67 & 80.72 & 81.52 & 87.74 & 75.59 & 25.12 \\
        $C^2$SDG~\cite{c2sdg}            
        & 73.11 & 84.47 & 98.59 & 99.33 & 83.30 & 13.32
        & 75.49 & 86.04 & 85.49 & 83.56 & 87.34 & 24.28 \\
        UNeXt-S~\cite{unext}              
        & 72.11 & 83.80 & 98.51 & 99.23 & 83.67 & 15.88
        & 74.41 & 85.33 & 85.00 & 84.71 & 85.27 & 25.17 \\
        MALUNet~\cite{malunet}            
        & 67.19 & 80.37 & 98.29 & 99.35 & 76.19 & 22.75
        & 74.64 & 85.48 & 85.70 & 89.29 & 82.27 & 24.17 \\
        EGE-UNet~\cite{egeunet}           
        & 65.81 & 79.38 & 98.35 & \textbf{99.77} & 68.95 & 19.29
        & 71.25 & 83.21 & 83.70 & 88.69 & 78.93 & 25.06 \\
        VM-UNet~\cite{vmunet}             
        & 72.02 & 83.74 & 98.37 & 98.72 & \textbf{91.11} & 15.29
        & 72.64 & 84.15 & 84.67 & 90.05 & 79.54 & 24.94 \\
        LightM-UNet~\cite{lightmunet}     
        & 71.44 & 83.34 & 98.59 & 99.67 & 76.32 & 15.37
        & 69.40 & 81.94 & 81.72 & 82.42 & 81.04 & 28.06 \\
        LB-UNet~\cite{lbunet}             
        & 63.75 & 77.86 & 98.14 & 99.45 & 71.07 & 14.49
        & 71.30 & 83.24 & 84.23 & \textbf{92.26} & 76.56 & 24.66 \\
        ULVM-UNet~\cite{ultralightvmnet}  
        & 70.19 & 82.49 & 98.49 & 99.53 & 76.97 & 15.23
        & 73.33 & 84.61 & 84.24 & 83.80 & 84.67 & 26.66 \\
        \textbf{Ours} 
        & \textbf{77.68} & \textbf{87.44} & \textbf{98.88} & 99.51 & 85.53 & \textbf{11.55}
        & \textbf{78.63} & \textbf{88.04} & \textbf{87.91} & 88.90 & 86.96 & \textbf{21.62} \\
        \hline
    \end{tabular}
    \caption{Performance comparison of SOTA models on BUS and GlaS datasets. For BUS, 163 samples are split into 114 for training, 18 for validation, and 31 reserved for testing. For GlaS, 165 samples are split into 70 for training and 15 for validation, with 80 reserved for testing. All models are trained and evaluated on the same partitions. (\(\uparrow\)) indicates higher is better. (\(\downarrow\)) indicates Lower is better. Best results are in bold.}
    \label{tab:bus_glas_generalization}
\end{table*}

\subsection{Robustness to Reduced Training Data}
To assess robustness under limited supervision, we train the model with only \{50, 70, 100\}\% of the original training split on ISIC2017 and ISIC2018, while keeping the test sets unchanged. As shown in Table~\ref{tab:data_eff_isic}, performance degrades steadily as the amount of training data decreases. On ISIC2017, reducing the training data from 100\% to 50\% lowers mIoU from 85.55 to 83.30 and DSC from 92.21 to 90.89, while HD95 increases from 10.73 to 13.24. A similar trend is observed on ISIC2018, where mIoU and DSC decrease from 83.60/91.07 to 81.49/89.80, while HD95 rises from 12.94 to 14.99. The performance drop remains modest on both datasets, which suggests that the model can still learn stable and discriminative representations even when annotation is substantially reduced. This behavior is especially important in medical image segmentation, where collecting dense pixel-level labels is costly and often limited.

\begin{table}[!t]
\centering
\footnotesize
\setlength{\tabcolsep}{4pt}
\renewcommand{\arraystretch}{1.05}
\caption{Robustness to reduced training data on ISIC2017 and ISIC2018. mIoU and DSC are reported in \%, and HD95 is reported in pixels. Results are obtained by randomly subsampling \{50, 70, 100\}\% of the training split, while keeping the test set unchanged. All metrics are computed on the full test set.}
\label{tab:data_eff_isic}
\begin{tabular}{c|ccc|ccc}
\hline
\multirow{2}{*}{\makecell{Training\\Data Size}}
& \multicolumn{3}{c|}{ISIC2017}
& \multicolumn{3}{c}{ISIC2018} \\
\cline{2-7}
& mIoU$\uparrow$ & DSC$\uparrow$ & HD95$\downarrow$
& mIoU$\uparrow$ & DSC$\uparrow$ & HD95$\downarrow$ \\
\hline
50\%  & 83.30 & 90.89 & 13.24 & 81.49 & 89.80 & 14.99 \\
70\%  & 84.14 & 91.39 & 12.06 & 82.23 & 90.25 & 13.61 \\
100\% & \textbf{85.55} & \textbf{92.21} & \textbf{10.73} & \textbf{83.60} & \textbf{91.07} & \textbf{12.94} \\
\hline
\end{tabular}
\end{table}

\section{Additional Ablation Study}
This section presents additional ablation studies to evaluate the impact of our design decisions further.

\subsection{Loss Function Analysis on PH2}
Table~\ref{tab:loss_ablation_ph2} compares three loss variants on the PH2 dataset \cite{ph2dataset}. Using binary cross-entropy (BCE) \cite{losssurvey} alone, we achieve 86.95\% IoU and 93.02\% DSC, while Dice loss (Dice) \cite{losssurvey} alone gives 86.59\% IoU and 92.81\% DSC. Combining BCE and Dice loss, MambaLiteUNet produces the best results (88.54\% IoU, 93.92\% DSC) and increases sensitivity to 90.45\%, representing an absolute sensitivity gain of 1.94 points over BCE loss alone and 2.88 points over Dice loss alone.
Therefore, our findings suggest that the hybrid loss stabilizes training and improves both overlap and boundary alignment.

\begin{table}[t]
    \centering
    \footnotesize
    \begin{tabular}{cc|ccccc}
        \hline
        \multicolumn{2}{c|}{Loss} 
            & \multicolumn{5}{c}{Performance Metrics} \\
            \hline
        BCE & Dice 
            & IoU$\uparrow$ & DSC$\uparrow$ & AC$\uparrow$ & SP$\uparrow$ & SE$\uparrow$ \\
        \hline
        \checkmark &            
            & 86.95 & 93.02 & 93.28 & 98.16 & 88.51 \\
                   & \checkmark  
            & 86.59 & 92.81 & 93.14 & \textbf{98.84} & 87.57 \\
        \checkmark & \checkmark 
            & \textbf{88.54} & \textbf{93.92} & \textbf{94.08} & 97.79 & \textbf{90.45} \\
        \hline
    \end{tabular}
    \caption{Results for different Loss functions on PH2. (\(\uparrow\)) indicates higher is better. $\checkmark$ indicates loss selection. Our results are averaged over five independent runs, and the best are in bold.}
    \label{tab:loss_ablation_ph2}
\end{table}

\begin{table*}[t!]
\centering
\footnotesize
\setlength{\tabcolsep}{4pt}
\renewcommand{\arraystretch}{1.2}
\begin{tabular}{ccc|cc|ccccc|ccccc}
\hline
\multicolumn{3}{c|}{Modules w/o Mamba} 
 & \multicolumn{2}{c|}{Complexity} 
 & \multicolumn{5}{c|}{ISIC2017} 
 & \multicolumn{5}{c}{ISIC2018} \\
\cline{1-15}
AMF & LGFM & CGA 
 & Params (M)$\downarrow$ & GFLOPs$\downarrow$ 
 & IoU$\uparrow$ & DSC$\uparrow$ & AC$\uparrow$ & SP$\uparrow$ & SE$\uparrow$
 & IoU$\uparrow$ & DSC$\uparrow$ & AC$\uparrow$ & SP$\uparrow$ & SE$\uparrow$ \\
\hline
\checkmark &           &           & 0.321 & \textbf{0.237} & 83.80 & 91.18 & 96.57 & 98.02 & 90.59 & 81.96 & 90.09 & 95.81 & \textbf{98.39} & 86.66 \\
           & \checkmark &           & \textbf{0.194} & 0.332 & 82.82 & 90.60 & 96.27 & 97.38 & \textbf{91.72} & 81.20 & 89.62 & 95.50 & 97.52 & 88.33 \\
           &           & \checkmark & 0.664 & 0.383 & 83.70 & 91.13 & 96.53 & 97.91 & 90.87 & 82.17 & 90.21 & 95.79 & 97.87 & 88.38 \\
\checkmark & \checkmark &           & 0.420 & 0.352 & 83.15 & 90.80 & 96.44 & 98.07 & 89.75 & 82.02 & 90.12 & 95.72 & 97.62 & 88.95 \\
\checkmark &           & \checkmark & 0.559 & 0.359 & 84.03 & 91.32 & \textbf{96.71} & \textbf{98.77} & 88.28 & 82.48 & 90.40 & 95.81 & 97.55 & \textbf{89.66} \\
           & \checkmark & \checkmark & 0.420 & 0.344 & 83.94 & 91.27 & 96.60 & 98.03 & 90.74 & 82.50 & 90.41 & 95.89 & 98.07 & 88.16 \\
\checkmark & \checkmark & \checkmark & 0.658 & 0.374 & \textbf{84.26} & \textbf{91.46} & 96.70 & 98.34 & 90.01 & \textbf{82.66} & \textbf{90.51} & \textbf{95.91} & 97.90 & 88.83 \\
\hline
\end{tabular}
\caption{Ablation study of AMF, LGFM, and CGA modules under the Mamba-off/Mamba-free control settings on ISIC2017 and ISIC2018 datasets. Complexity is measured by Params (M) and GFLOPs. (\(\uparrow\)) indicates higher is better, while (\(\downarrow\)) indicates lower is better.}
\label{tab:module_contributions}
\end{table*}

\begin{table*}[t]
\centering
\footnotesize
\renewcommand{\arraystretch}{1.2}
\setlength{\tabcolsep}{6pt} 
\begin{tabular}{p{1cm} p{3.1cm} p{4.2cm} p{2.2cm} p{4.2cm}}
\toprule
Module & Design Goal & Our Mechanism & Closest Prior & Key Distinction $\rightarrow$ Expected Benefit \\
\midrule
AMF &
Scale–adaptive multi-branch feature fusion under tight compute. &
Channels are split into parallel Mamba SSM branches, then merged through a two-stage DW$\rightarrow$PW gating pipeline that adapts routing to the input. Residual reweighting ensures stability. &
ResNeXt (grouped conv with fixed cardinality). &
Fixed partitions in ResNeXt vs. dynamic gating with Mamba branches + dual gates
$\rightarrow$ Content-aware allocation of capacity, sharper interiors, and reduced under-/over-emphasis structures at similar cost. \\
\hline

LGFM &
Fuse local detail with long-range context inside a single block. &
A dual-path block: DWConv($3{\times}3$) for textures and edges, + MHA for global dependencies. Features are concatenated and projected back in a single residual unit (no external fusion head). &
TransFuse (CNN and Transformer encoders fused by external BiFusion). & Separate encoders + late fusion vs.
In-block local–global mixing.
$\rightarrow$ Less redundancy, balanced paths, and stronger boundary retention (lower HD95). \\
\hline

CGA &
Denoise or reduce background information and regulate skip connections before decoder fusion. &
Cross-gated skip aggregation: encoder–decoder pairs refined with Mamba, projected through DWConv+sigmoid, and gated bidirectionally before fusion. &
Attention U-Net (decoder-driven, one-way gating of encoder skips). &
One-way decoder gating vs. Bidirectional pre-fusion gating.
$\rightarrow$ Cleaner skips, suppressed background noises, and sharpens edges with minimal overhead. \\

\bottomrule
\end{tabular}
\caption{Comparative analysis of our AMF, LGFM, and CGA modules against their closest priors (ResNeXt~\cite{resnext}, TransFuse~\cite{transfuse}, and Attention U-Net~\cite{attentionunet}). The table highlights how architectural differences in design goals and mechanisms translate into measurable segmentation gains. Abbreviations: DW: depthwise convolution; PW: pointwise convolution; SSM: state space model~\cite{mambaoriginal}; MHA: multi-head self-attention~\cite{vit}.}
\label{tab:comparative_modules}
\end{table*}

\subsection{Effect of Core Architectural Modules without (w/o) Mamba Integration}

To isolate the contributions of our Adaptive Multi-branch Feature Fusion (AMF), Local-Global Feature Mixing (LGFM), and Cross-Gated Attention (CGA) from SSM-based long-range modeling, we design Mamba-off control experiments. In this configuration, all Mamba layers are replaced with token-MLPs (two-layer feed-forward networks with LayerNorm, GELU nonlinearity, and residual connection). In this substitution, we aim to preserve a similar parameter count and the same channel dimensionality while removing Mamba's structured recurrent dynamics. 

Our originally proposed AMF, LGFM, and CGA modules are inspired by selective gating principles and incorporate Mamba layers in the full model. In the Mamba-off control, however, these modules adopt the same selective gating principles but are implemented using convolutional layers, multi-head self-attention, and token-MLPs, without invoking Mamba kernels or SSM recurrence. Consequently, we can disentangle the contribution of the Mamba-integrated and Mamba-off approaches.   

Table~\ref{tab:module_contributions} shows the impact of AMF, LGFM, and CGA under the Mamba-off setting.  On ISIC2017, AMF alone achieves the best single-module improvement (83.80\% IoU, 91.18\% DSC) with minimal cost, while LGFM improves sensitivity (91.72\%). CGA provides balanced gains but requires higher complexity. Pairwise combinations further enhance performance, with AMF+CGA achieving the highest accuracy and specificity, and the full configuration reaching the best overall results (84.26\% IoU, 91.46\% DSC). On ISIC2018, results trends are consistent: AMF improves overlap, LGFM boosts sensitivity, and CGA strengthens boundary quality. The full setup (AMF+LGFM+CGA without Mamba) achieves 82.66\% IoU and 90.51\% DSC, with 0.658M parameters and 0.374 GFLOPs. 

However, when comparing with the Mamba-integrated configuration (see Table~7 in the main manuscript), we observe clear performance boosts. With Mamba, individual modules improve their performance. The full design (AMF+LGFM+CGA with Mamba) achieves 85.55\% IoU and 92.21\% DSC on ISIC2017 and 83.60\% IoU and 91.07\% DSC on ISIC2018, outperforming all Mamba-off results while remaining lightweight (0.494M parameters, 0.326 GFLOPs). Therefore, Mamba integration with AMF, LGFM, and CGA is critical, which demonstrates consistent improvements across datasets with minimal overhead.

\section{Comparative Analysis of Module Designs}
\label{sec:s6_comparative_modules}
To further clarify the novelty of our proposed AMF, LGFM, and CGA, we provide a detailed comparison with their closest prior designs. Table~\ref{tab:comparative_modules} summarizes each module’s design goal, mechanism, nearest prior, and the architectural differences that lead to the expected improvements in lesion segmentation.

\section{Module-wise Feature Map Visualization}
\begin{figure}[t]
    \centering
    \includegraphics[width=0.98\linewidth]{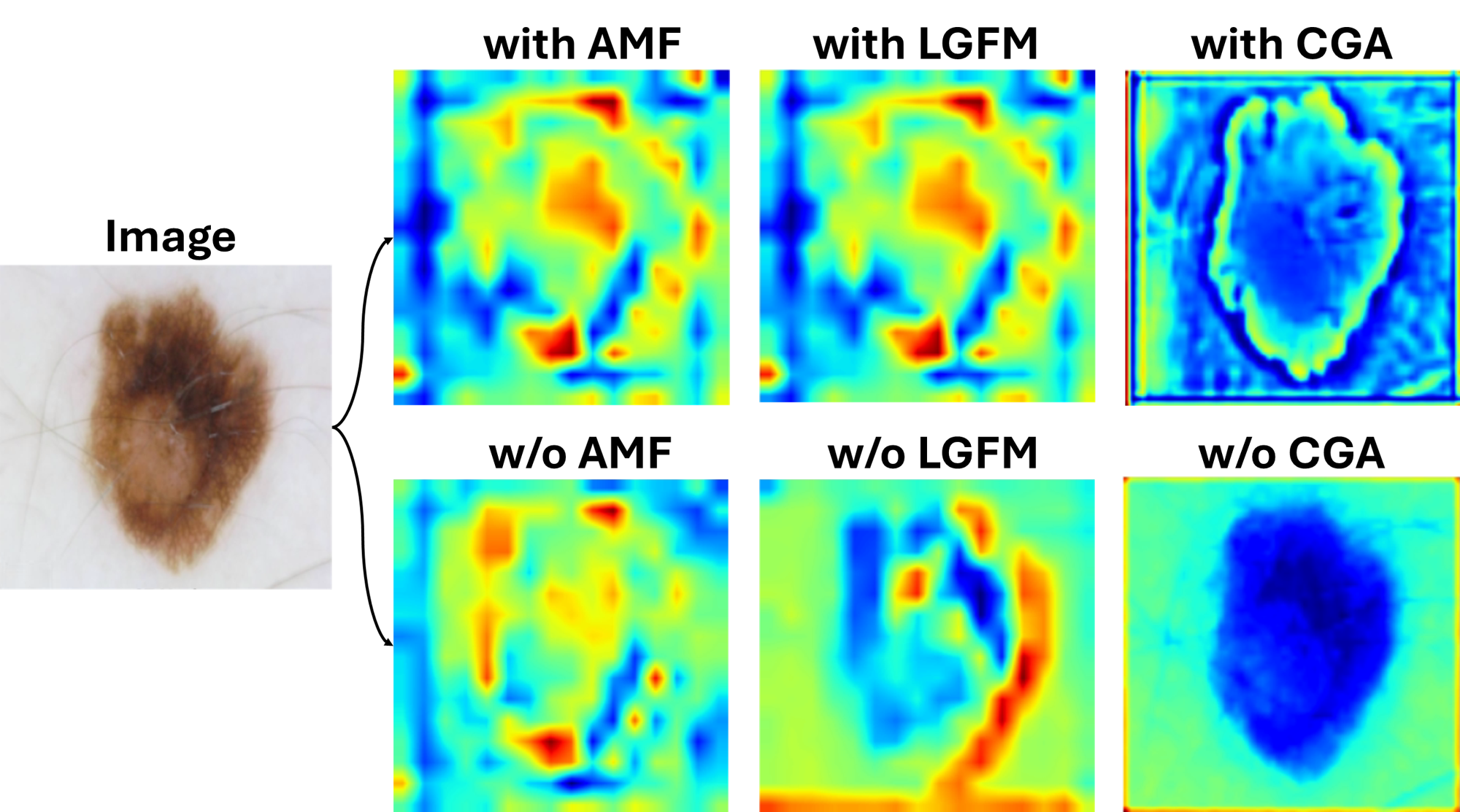}
    \caption{Qualitative visualization of module contributions. The top row shows representative feature maps when each module is enabled, while the bottom row shows the corresponding maps when the module is removed. “with” denotes the full model containing the module, and “w/o” denotes the variant where the module is disabled.}
    \label{fig:module_vis}
\end{figure}

Figure~\ref{fig:module_vis} provides a qualitative comparison of representative feature maps with and without the key modules in MambaLiteUNet. The top row presents the feature responses from the full model with AMF, LGFM, and CGA, while the bottom row shows the corresponding feature responses after removing each module. This comparison highlights how each component shapes the internal spatial representation.

With AMF, the feature response is more structured and lesion-aware, which reflects its role in adaptive feature refinement. LGFM produces the clearest and most coherent lesion-focused activation; when it is removed, the feature map becomes weaker and less discriminative, indicating the importance of local-global feature integration. CGA mainly strengthens boundary-sensitive structure. With CGA, the lesion contour is more clearly emphasized, whereas removing it yields a smoother and less selective response around the lesion region.
Therefore, these visualizations show that the three modules contribute in complementary ways. AMF improves adaptive refinement, LGFM strengthens lesion-focused representation, and CGA enhances boundary-aware filtering. Together, they produce more informative and spatially coherent intermediate features.

\section{Stage-wise Feature Map Visualization}

\begin{figure*}[t]
    \centering
    \includegraphics[width=\textwidth]{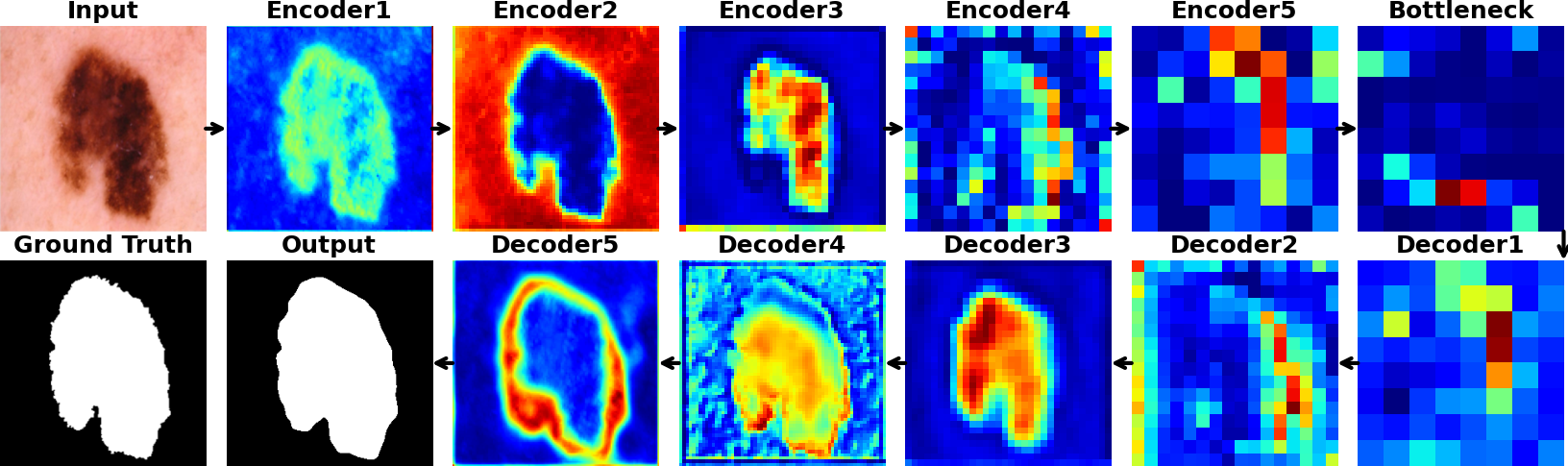}
    \caption{Feature map visualization in MambaLiteUNet. Top row (left→right): input image, Encoder1–Encoder5, Bottleneck—features grow more abstract and emphasize lesion boundaries. Bottom row (left→right): ground‑truth mask, model output, and Decoder5–Decoder1 activations. Decoding flows from Decoder1 through Decoder5 to the output (arrows): early blocks (Decoder1/Decoder2) recover coarse structure, while later blocks (Decoder4/Decoder5) refine and sharpen lesion contours. Arrows indicate the forward flow of information.}
    \label{fig:feature_maps}
\end{figure*}

This section provides stage-wise qualitative evidence of how MambaLiteUNet processes lesion images throughout its encoder--decoder pipeline, complementing the quantitative results presented in the main manuscript. Figure~\ref{fig:feature_maps} illustrates how our proposed MambaLiteUNet progressively transforms feature representations. We visualize intermediate feature maps using a model pre‑trained on ISIC2018 \cite{isic2018challenge} and tested on a held‑out image. The top row shows the input image, followed by activation maps from each encoder stage (Encoder1–Encoder5) and the bottleneck. As depth increases, the model learns progressively more abstract and localized features that emphasize lesion boundaries and suppress background noise.

The bottom row (left→right) shows the ground‑truth mask, the model’s final output, and then decoder activations from Decoder5 through Decoder1. Early decoder blocks (Decoder1 and Decoder2) recover coarse structure, while later blocks (Decoder3–Decoder5) refine contours and sharpen lesion boundaries. This validates our model’s hierarchical encoding, skip‑guided decoding, and reconstruction. Arrows indicate the forward flow of information through the network.

\begin{table}[h]
\centering
\footnotesize
\setlength{\tabcolsep}{3mm}
\renewcommand{\arraystretch}{1.12}
\begin{tabular}{l|c|c}
\hline
Model & Sec/Image $\downarrow$ & Memory (MB) $\downarrow$ \\
\hline
VM-UNet~\cite{vmunet}             & 0.1718 &  582.5 \\ 
VM-UNet2~\cite{vmunet2}           & 0.1836 &  613.7 \\
LightM-UNet~\cite{lightmunet}     & 0.0194 &   63.6 \\
ULVM-UNet~\cite{ultralightvmnet}  & \textbf{0.0058} & \textbf{17.4} \\
\textbf{Ours}                     & 0.0167 & 54.5 \\
\hline
\end{tabular}
\caption{Inference efficiency comparison among Mamba-based models. Latency (Sec/Image) and peak GPU memory (MB) at $256\times256$ with batch size $1$ evaluated on an NVIDIA RTX 3090 Ti (24 GB) using CUDA timing and PyTorch memory profiling. ($\downarrow$) indicates lower is better. Best results in bold.}
\label{tab:inference_and_memory}
\end{table}

\section{Inference Time and Memory Usage}
Table~\ref{tab:inference_and_memory} presents a comparative analysis of inference time and memory for Mamba-based models. VM-UNet (0.1718 Sec/Image, 582.5 MB) and VM-UNet2 (0.1836 Sec/Image, 613.7 MB) are the most computationally expensive. LightM-UNet is considerably lighter (0.0194 Sec/Image, 63.6 MB), and ULVM-UNet achieves the best efficiency at 0.0058 Sec/Image and 17.4 MB. In comparison, our MambaLiteUNet operates at 0.0167 Sec/Image with 54.5 MB, slightly above ULVM-UNet in cost but offering stronger representational power and higher segmentation accuracy (see Sec.~\ref{supp:add_exp_results}), making it a balanced choice for accuracy and deployment efficiency.


\end{document}